\documentclass[journal]{IEEEtran}
\usepackage[ruled]{algorithm2e}
\usepackage{algorithmic}
\usepackage{color,xcolor,soul}
\usepackage{graphicx}
\usepackage{epstopdf}
\usepackage{mhchem}
\usepackage{subfigure}
\usepackage{bm}
\usepackage{tabularx}
\usepackage{colortbl,booktabs}
\usepackage{multirow}
\usepackage{adjustbox}
\usepackage{array}
\usepackage{makecell}
\usepackage[colorlinks,linkcolor=blue]{hyperref}

\newcommand{\PreserveBackslash}[1]{\let\temp=\\#1\let\\=\temp}
\newcolumntype{C}[1]{>{\PreserveBackslash\centering}p{#1}}
\newcolumntype{R}[1]{>{\PreserveBackslash\raggedleft}p{#1}}
\newcolumntype{L}[1]{>{\PreserveBackslash\raggedright}p{#1}}

\begin{document}
	\title{An Efficient and Adaptive Granular-ball Generation Method in Classification Problem}

	\author{Shuyin~Xia,
		Xiaochuan~Dai,
		Guoyin~Wang*,~\IEEEmembership{Senior~Member,~IEEE,}
		Xinbo Gao,~\IEEEmembership{Senior~Member,~IEEE,}
		Elisabeth Giem
		\IEEEcompsocitemizethanks{\IEEEcompsocthanksitem S. Xia, X. Dai, G. Wang\& X. Gao are with the Chongqing Key Laboratory of Computational Intelligence, Chongqing University of Telecommunications and Posts, 400065, Chongqing, China. E-mail: xiasy@cqupt.edu.cn, daixiaochuaner@qq.com, wanggy@cqupt.edu.cn, gaoxb@cqupt.edu.cn. \protect\\			
			\IEEEcompsocthanksitem E. Giem is with the Department of Computer Science and Engineering, University of California Riverside, Riverside, CA, 92521. E-mail: gieme01@ucr.edu}}
	\maketitle
	
	\begin{abstract}
		Granular-ball computing is an efficient, robust, and scalable learning method for granular computing. The basis of granular-ball computing is the granular-ball generation method. This paper proposes a method for accelerating the granular-ball generation using the division to replace $k$-means. It can greatly improve the efficiency of granular-ball generation while ensuring the accuracy similar to the existing method. Besides, a new adaptive method for the granular-ball generation is proposed by considering granular-ball's overlap eliminating and some other factors. This makes the granular-ball generation process of  parameter-free and completely adaptive in the true sense. In addition, this paper first provides the mathematical models for the granular-ball covering. The experimental results on some real data sets demonstrate that the proposed two granular-ball generation methods have similar accuracies with the existing method while adaptiveness or acceleration is realized. All codes have been released in the open source GBC library at http://www.cquptshuyinxia.com/GBC.html.
	\end{abstract}
	

	\IEEEpeerreviewmaketitle

	\vspace{-1em}	
	\section{Introduction}
	
\begin{figure}[!t]
		\centering
		\includegraphics[width=0.25\textwidth]{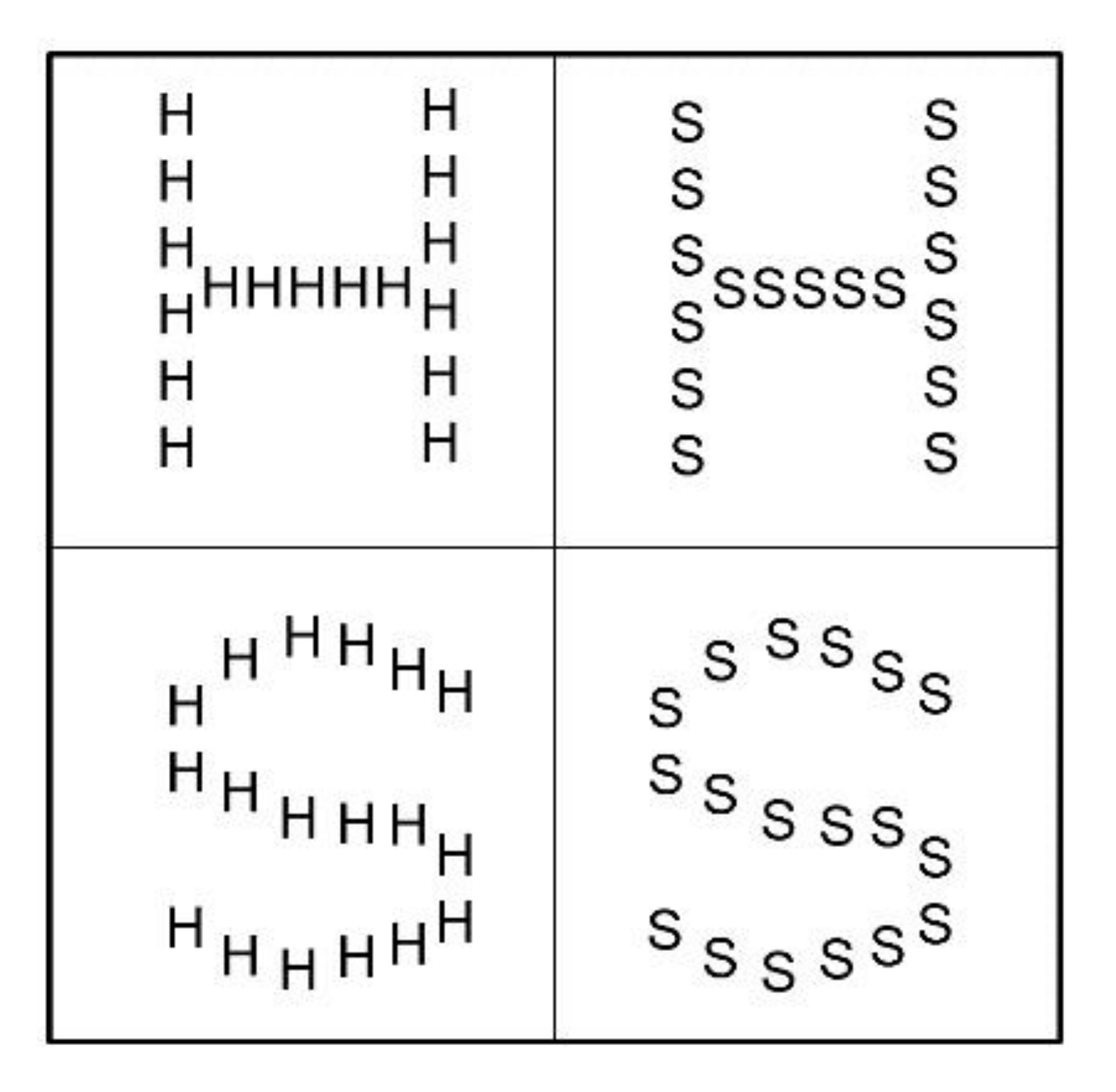}
		\caption{Human cognition - coarse grained large range is preferred.}
		\label{Fig1} 
	\end{figure}

	\IEEEPARstart{C}{OGNITIVE} computing combined with human cognitive mechanism makes the decision-making process more reliable, efficient and understandable. It is an important means to achieve reliable governance of information space and an important direction for the development of artificial intelligence. Academician Chen's research results published in sciences in 1982 pointed out that human cognition is characterized by large-scale priority~\cite{chen1982topological}. As shown in Fig. \ref{Fig1}, the large outline letters are seen first, followed by the specific small letters within the outline letters. Based on this cognitive characteristic, granular computing can achieve efficient, scalable and robust learning processes. Zadeh, a famous American cybernetics expert, put forward the problem of granular information granulation and the concept of granular computing~\cite{zadeh1979fuzzy, zadeh1997toward}. After decades of continuous research by scholars at home and abroad, fuzzy sets~\cite{4backer1981clustering, 5kosko1986counting, 6zhang2020multi, 7liu2019combinatorial} and rough sets~\cite{8xia2020gbnrs, 9liang2012group, 10qian2010positive, 11hu2017large} have been developed. Academician Bo Zhang proposed the quotient space theory~\cite{12zhang2004quotient, 13ling2003theory}, academician Deyi Li proposed the cloud model theory~\cite{14li1998uncertainty, 15li2009new}, and Guoyin Wang and Shuyin Xia developed the granular-ball computing~\cite{16xia2020fast, 17xia2019granular} and other model methods. 
	
	In fuzzy sets, data elements are described by membership degree of different granularity of fuzzy information. Rough set theory and quotient space theory use equivalence relations and equivalence classes to construct granules of different sizes. The universe of rough set theory is the point set of objects, and the topological relations between elements are not considered. The quotient space theory is studied on the condition that there are topological relations between the elements in the universe. The quotient space and cloud model are also two important granular computing methods. Among them, rough sets have been the most widely studied, and many scholars have made a lot of outstanding work in the field of rough sets. Guoyin Wang found the difference between algebraic form and information entropy form of rough set~\cite{18wang2003rough}. Duoqian Miao and others found the equivalence between fuzzy soft sets and fuzzy information systems~\cite{19pei2005soft}. Yuhua Qian, Jiye Liang and others reduced the amount of calculations in the positive field in the attribute combination explosion through positive field reduction and nuclear attributes~\cite{10qian2010positive}. Zeshui Xu and others established the topological structure of covering rough set model~\cite{20xu2005properties}. Yiyu Yao pointed out that label noise had an obvious interference effect on upper and lower approximate calculation~\cite{21yao2007decision}. Qinghua Hu and others designed a robust classification algorithm based on fuzzy lower approximation~\cite{22hu2011robust}, and consider the data distribution information and be included in the calculation and fuzzy approximate the data distribution of rough set model perception~\cite{23an2015data}. Weizhi Wu and others proposed a multi-scale decision table~\cite{24wu2011theory}. Tianrui Li put forward the dynamic incremental approximation method based on rough set maintenance environment~\cite{25chen2011rough}. Shuyin Xia and Guoyin Wang proposed a parameter-less rough set method that could process continuous data without relying on membership function~\cite{8xia2020gbnrs}.
	
\begin{figure}[!t]
		\centering
		\subfigure[]{\includegraphics[width=0.23\textwidth]{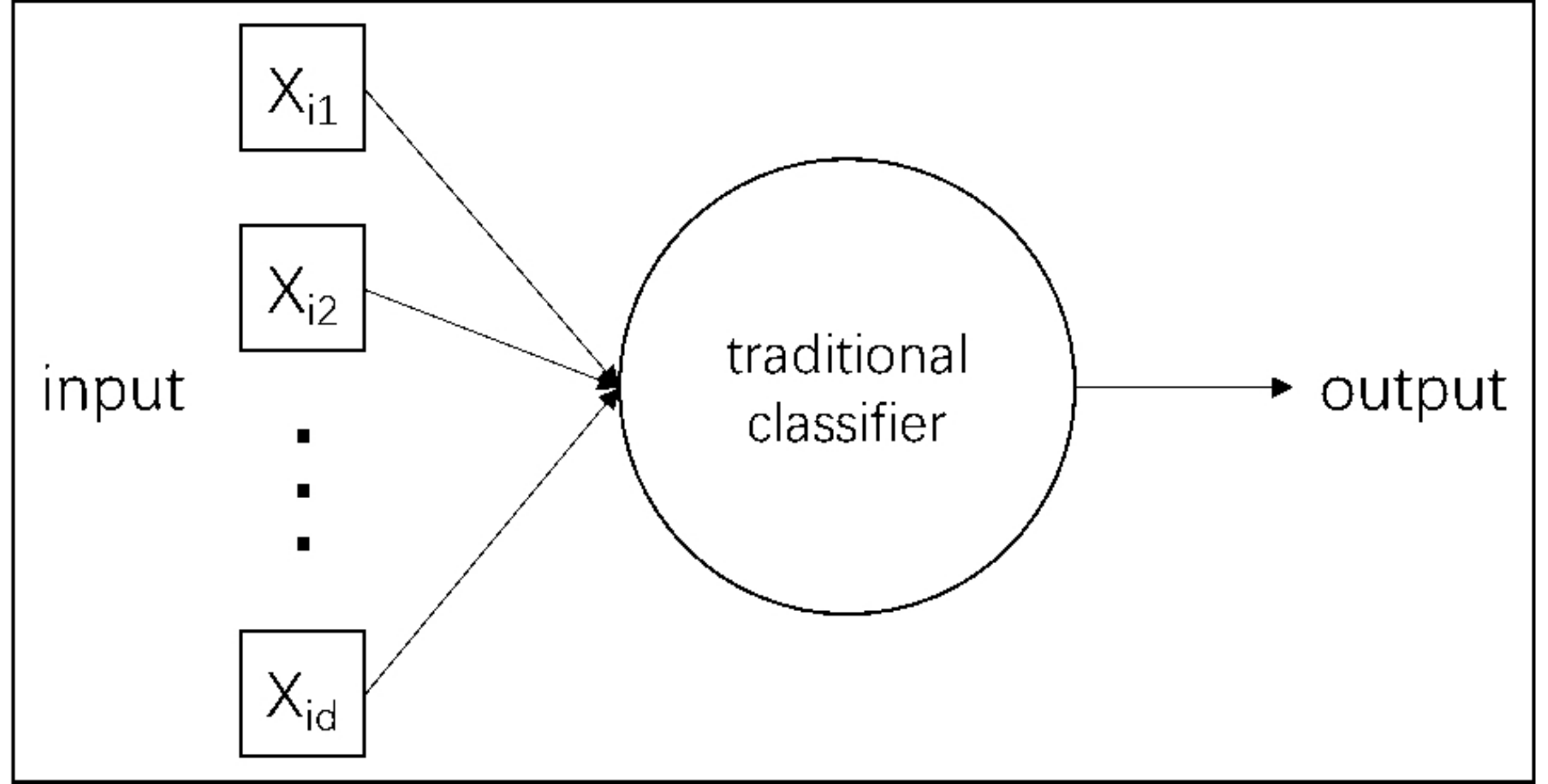}}
		\subfigure[]{\includegraphics[width=0.23\textwidth]{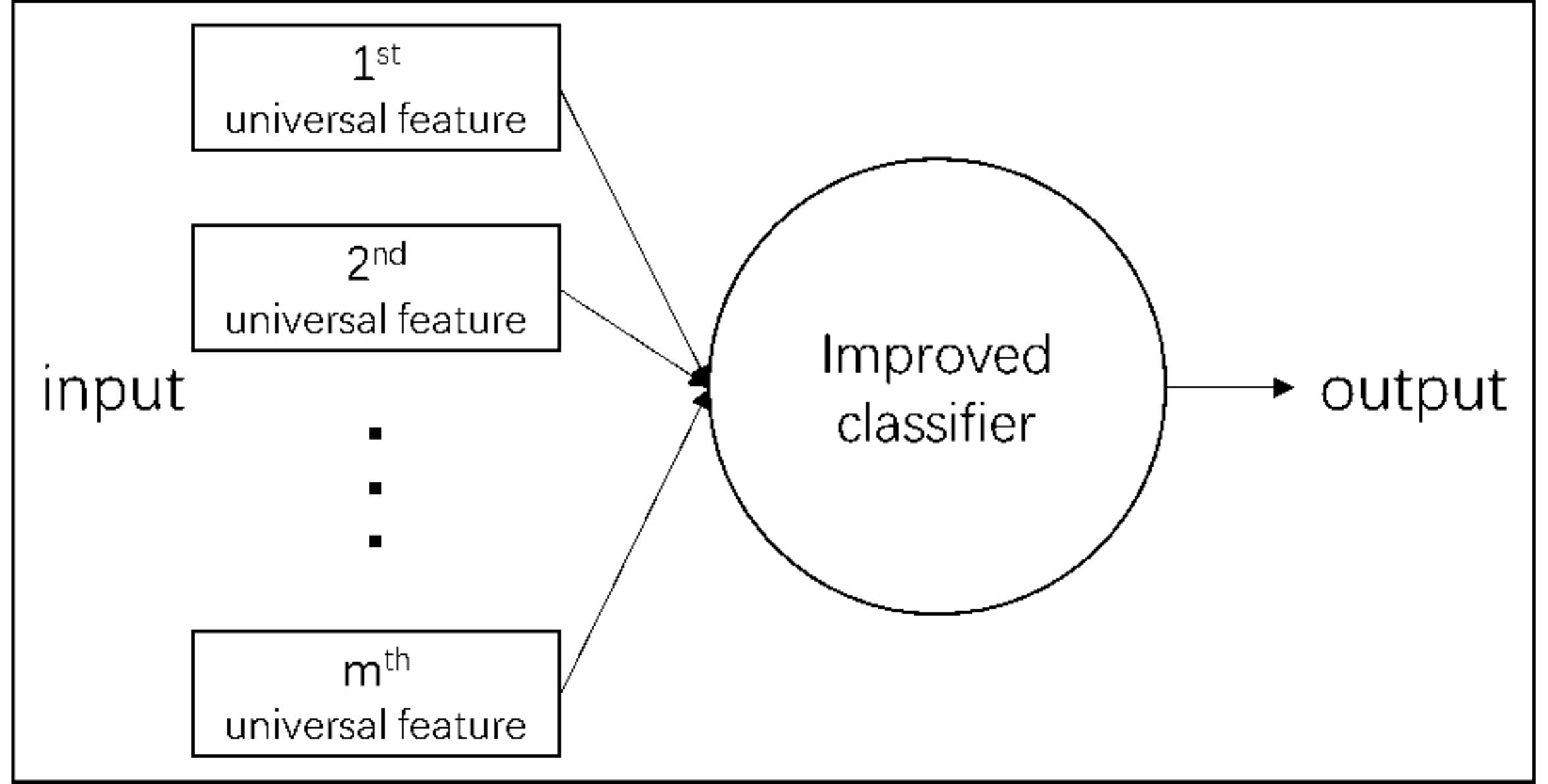}}
		\centering{\caption{Comparison between multi granularity classifier and traditional classifier. (a) Traditional classifier method; (b) Classifier method of coarse-grained input.}
		\label{Fig2}}
		\vspace{-1em}
	\end{figure}

	In granular computing, the larger the granularity, the higher the efficiency, and the better the robustness to noise; but it is also more likely to cause neglect of details and loss of accuracy. The smaller the granularity is, the more attention is paid to the details, but it may reduce efficiency and deteriorate robustness to noise. Selecting different granularity according to different scenes can better play the performance of multi-granularity learning method. Although multi-granular computing has a long research history, as a cognitive computing science, it also faces some new challenges and needs new development. For example, in terms of the ``classifier", one of the most widely used methods of artificial intelligence, as shown in Fig. \ref{Fig2}(a), the input of most existing classifiers is the finest-grained sample points or pixels~\cite{26salehi2015synergistic, 27cover1967nearest, 28loh2011classification}, , so there will be a lack of coarse-grained characterization. Some researchers have proposed classification algorithms based on multi-granularity ideas. For example, Dick S and others assumed that connection weights are all linguistic variables, and they have different granulation of connection weights~\cite{29dick2001granular}. Weight updating is realized by adding ``linguistic hedges", but this method will sacrifice certain accuracy. Leite D, MM. M. Gupta and FY Wang and others used fuzzy neurons to build interpretable multi-size local models~\cite{30leite2013evolving, 31gupta1990fuzzy, 32wang1995implementing}, which can learn fuzzy rules, and the output space can be processed as membership information, which can be used to process fuzzy data. In a few cases for machine learning and data mining, such as~\cite{33syeda2002parallel}, fuzzy rules are extracted for credit card fraud detection. The purpose of these works is to use neural networks to process fuzzy data and apply it to fuzzy control. It is not based on multi-granularity ideas to improve the scalability, efficiency, or robustness of the classifier, and its essence is still a point input method. Park HS and others constructed the information granule by feature selection in the input space, which is essentially a feature preprocessing method without changing the learning mode of the neural networks~\cite{34park2009granular}. Tang Y and others introduced a method of sampling and mapping information granules in a support vector machine~\cite{35tang2012granular}. The research work of ~\cite{34park2009granular, 35tang2012granular} focuses more on understanding some existing research work by using the concept of multi-granularity. Pedrycz W and others systematically proposed a neural network granulation framework from the input layer and output layers~\cite{36pedrycz2001granular}. Both rough set and fuzzy methods can granulate the input space. But this work does not realize a specific multi-granularity neural network and examine its performance advantages. Therefore, how to implement the multi-granularity classifier shown in Fig. \ref{Fig2}(b) is an important challenge. In the multi-granularity classifier in Fig. \ref{Fig2}(b), the input is no longer the finest-grained point, but a universal feature with adjustable granularity. The design of this universal feature should meet high-dimensional scalability, that is, no complicated calculations are required in the high-dimensional space, otherwise the high-dimensional problem cannot be dealt with. For this reason, Guoyin Wang and Shuyin Xia proposed the use of ball as ``granule" to represent this universal feature and proposed a granular-ball computing method~\cite{17xia2019granular}. The reason is that the geometry of a ball is completely symmetrical, and only two data are needed to characterize it in any dimension: center and radius, so it is convenient to be applied to high-dimensional data. At the same time, they also proposed an efficient and adaptive method to generate granular-balls. 
	
	Further, the granular-ball computing is introduced into the classifier, the framework of the granular-ball computing classifier is proposed, and the original model of the granular-ball supports vector machine (GBSVM) is derived and the $k$-nearest neighbor algorithm of the granular-ball (GB$k$NN) is proposed~\cite{17xia2019granular}. The efficiency of GB$k$NN is hundreds of times higher than that of the existing $k$NN algorithm, especially in large-scale data. In addition, GB$k$NN does not need to select parameter $k$, and helps to alleviate the performance in unbalanced data, which is not available in existing $k$-nearest neighbor algorithms. Due to the robustness of granular-ball computing, GB$k$NN has higher accuracy than accurate $k$NN in many data. In addition, granular-ball computing was introduced into the neighborhood rough set, and a new rough set method ``granular-balls neighborhood rough set (GBNRS)" was developed~\cite{8xia2020gbnrs}. GBNRS is the first parameter-free rough set algorithm to process continuous data without prior knowledge (i.e. setting membership function), which is more efficient than NRS. Since GBNRS can adaptively select the neighborhood radius, it can also obtain higher classification accuracy than NRS in many cases. In addition, the granular-ball computing was introduced into the $k$-means algorithm and a simple and fast $k$-means clustering method ``ball $k$-means" is developed~\cite{16xia2020fast}. Ball $k$-means is dozens of times more efficient than similar algorithms, especially in the challenging large-$k$ clustering problem. Granular-ball computing is efficient, robust and scalable~\cite{17xia2019granular}. However, there are still many challenges in the granular-ball generation, such as the optimization of the purity threshold and its efficiency improvement. The main contributions of this paper are as follows:	
	\begin{enumerate}[\IEEEsetlabelwidth{4)}]
		\item The acceleration granular-ball generation method is proposed using the division to replace $k$-means. It can accelerate the granular-ball generation several times to dozens of times while a similar accuracy is achieved.
		
		\item A new adaptive method for the granular-ball generation is proposed by considering granular-ball's overlap eliminating and some other factors. This makes the granular-ball generation process of parameter-free and completely adaptive in the true sense.
		
		\item This paper first provides the mathematical models for the granular-ball covering. 
	\end{enumerate}	
	
\section{Related Work}
	
	\subsection{Granular-ball Computing}
	
\begin{figure}[htbp!]
		\centering
		\fbox{\includegraphics[scale=0.25]{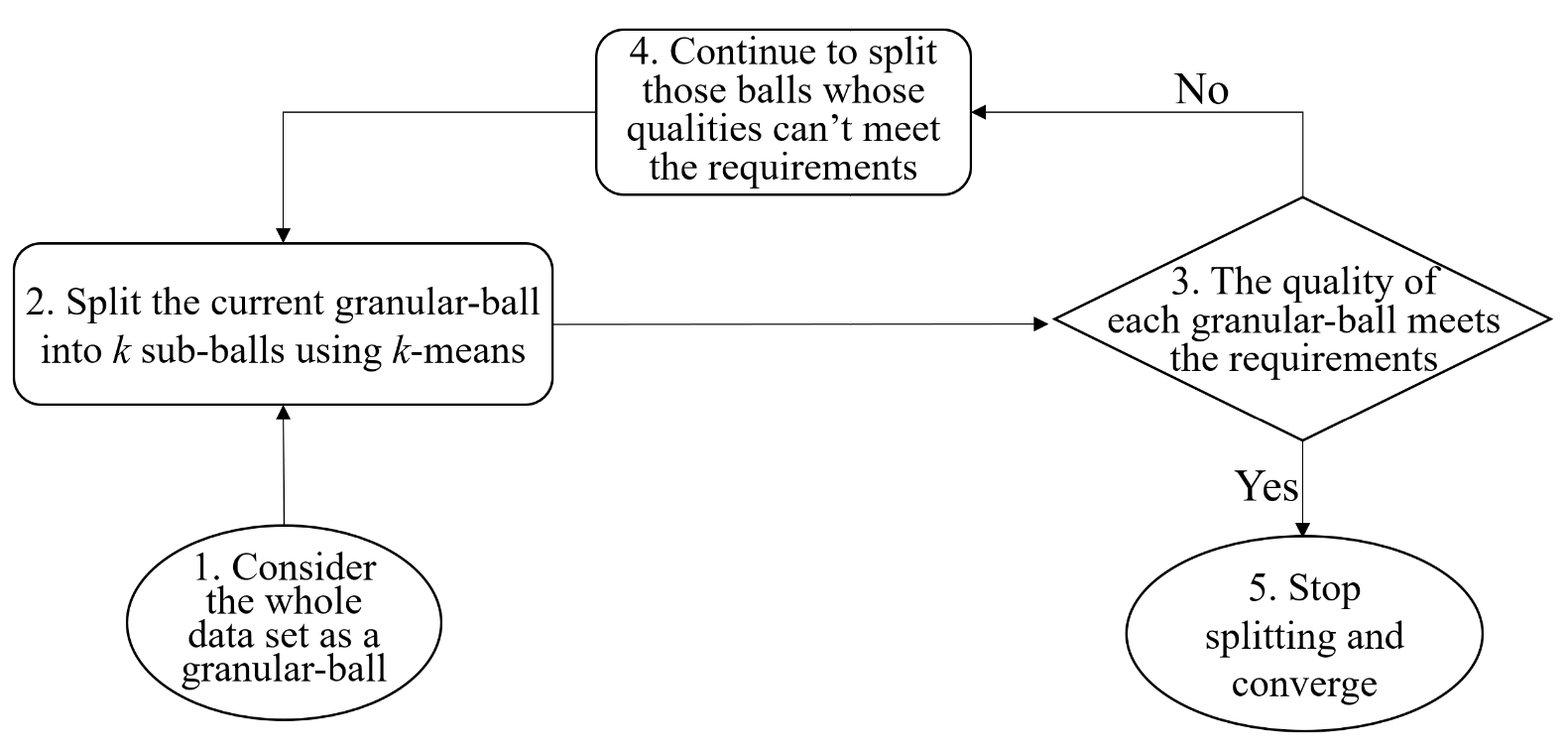}}
		\caption{Process of the existing granular-ball generation in granular-ball computing.} 
		\label{fig:GBCProcess}	
	\end{figure}

	Combining the theoretical basis of traditional granular computing, and based on the research results published by Chen in Science in 1982, he pointed out that ``human cognition has the characteristics of large-scale priority" ~\cite{chen1982topological}, Wang put forward a lot of granular cognitive computing~\cite{56}. Based on granular cognitive computing, granular-ball computing is a new, efficient and robust granular computing method proposed by Xia and Wang~\cite{17xia2019granular}, the core idea of which is to use ``granular-balls" to cover or partially cover the sample space. A granular-ball $GB=\{x_i,i=1...N\}$, where $x_i$ represents the objects in $GB$, and $N$ is the number of objects in $GB$. $GB$'s center $C$ and radius $r$ are respectively represented as follows
	\begin{equation} \label{equ:center}
		C=\frac {1}{N}{}\sum\limits_{i=1}^{N}{x_{i}},
	\end{equation}
	\begin{equation}
		r=\frac{1}{N}\sum\limits_{i=1}^{N}{\left| {{x}_{i}}-C \right|}.
	\end{equation}
	This means that the radius is equal to the average distance from all objects in $GB$ to its center. The radius can also be set to the maximum distance. The ``granular-ball" with a center and radius are used as the input of the learning method or as accurate measurements to represent the sample space, achieving multi-granularity learning characteristics (that is, scalability, multiple scales, etc.) and the accurate characterization of the sample space. The basic process of granular-ball generation for classification problems in granular-ball computing is shown in Fig. \ref{fig:GBCProcess}.
	
	As shown in Fig. \ref{fig:GBCProcess}, to simulate the ``the characteristics of large-scale priority of human cognition" at the beginning of the algorithm, the whole data set can be regarded as a granular-ball. At this time, the purity of the granular-ball is the worst and cannot describe any distribution characteristics of the data. The ``purity" is used to measure the quality of a granular-ball ~\cite{17xia2019granular} in the step 3 in Fig. \ref{fig:GBCProcess}. It is equal to the proportion of the most labels in the granular-ball. Then, the number of different classes in the granular-ball is counted and denoted as $m$; the granular-ball is split into $m$ child granular-balls in the step 2. In the step 3, the purity of each granular-ball is calculated; if a granular-ball does not reaches the purity threshold, it needs to be split. As the splitting process continues to advance, the purity of the granular-balls increases, and the decision boundary becomes increasingly clearer; the boundary is clearest, and the algorithm converges when the purity of all granular-balls meets the requirements. It can be concluded from ~\cite{17xia2019granular} that for a data set, no matter what distribution its data has, we can describe its decision boundary by enough granular-balls. 
	
	\begin{figure}[htbp!]
		\centering
		\subfigure[]{\includegraphics[width=0.23\textwidth]{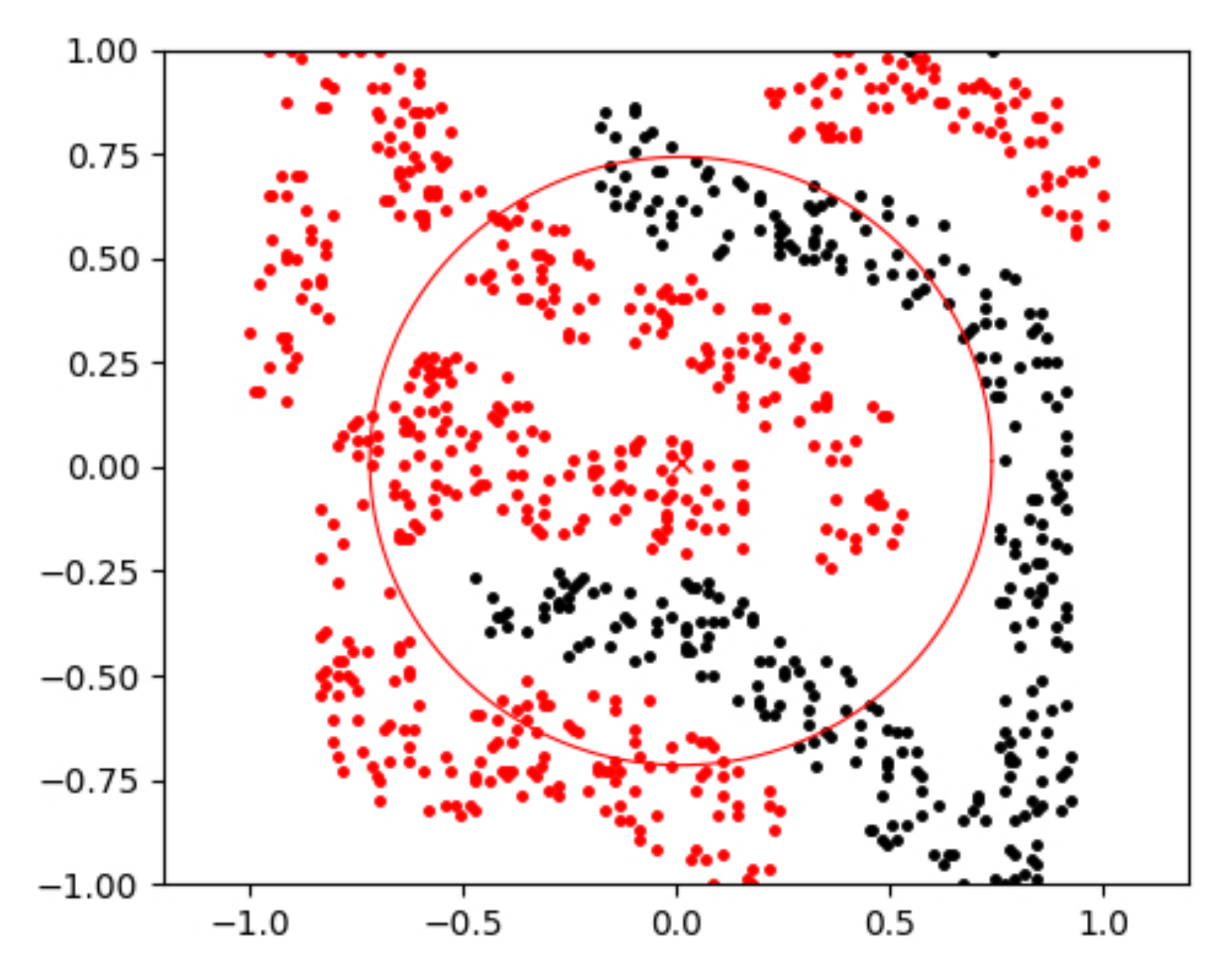}}
		\subfigure[]{\includegraphics[width=0.23\textwidth]{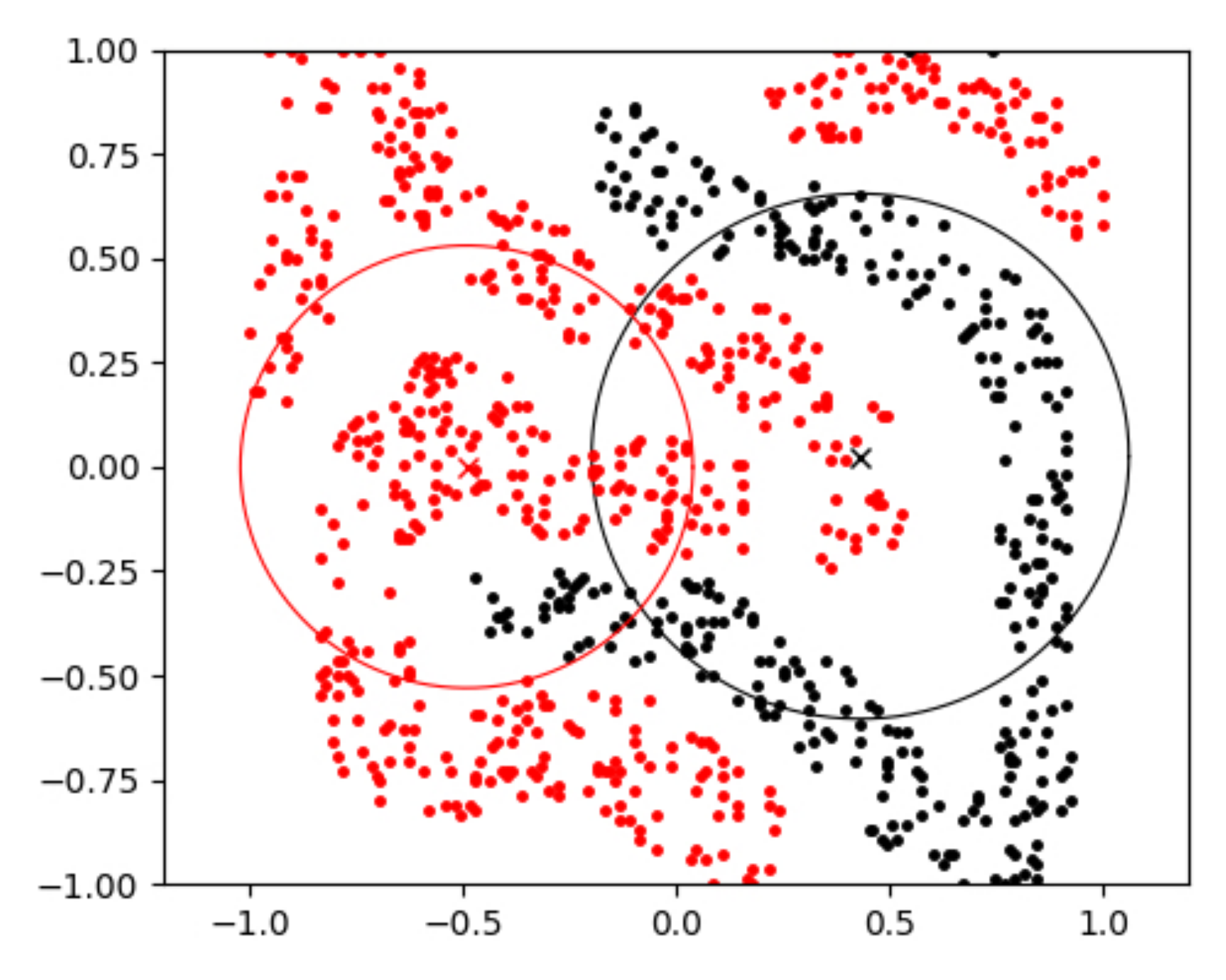}}
		\subfigure[]{\includegraphics[width=0.23\textwidth]{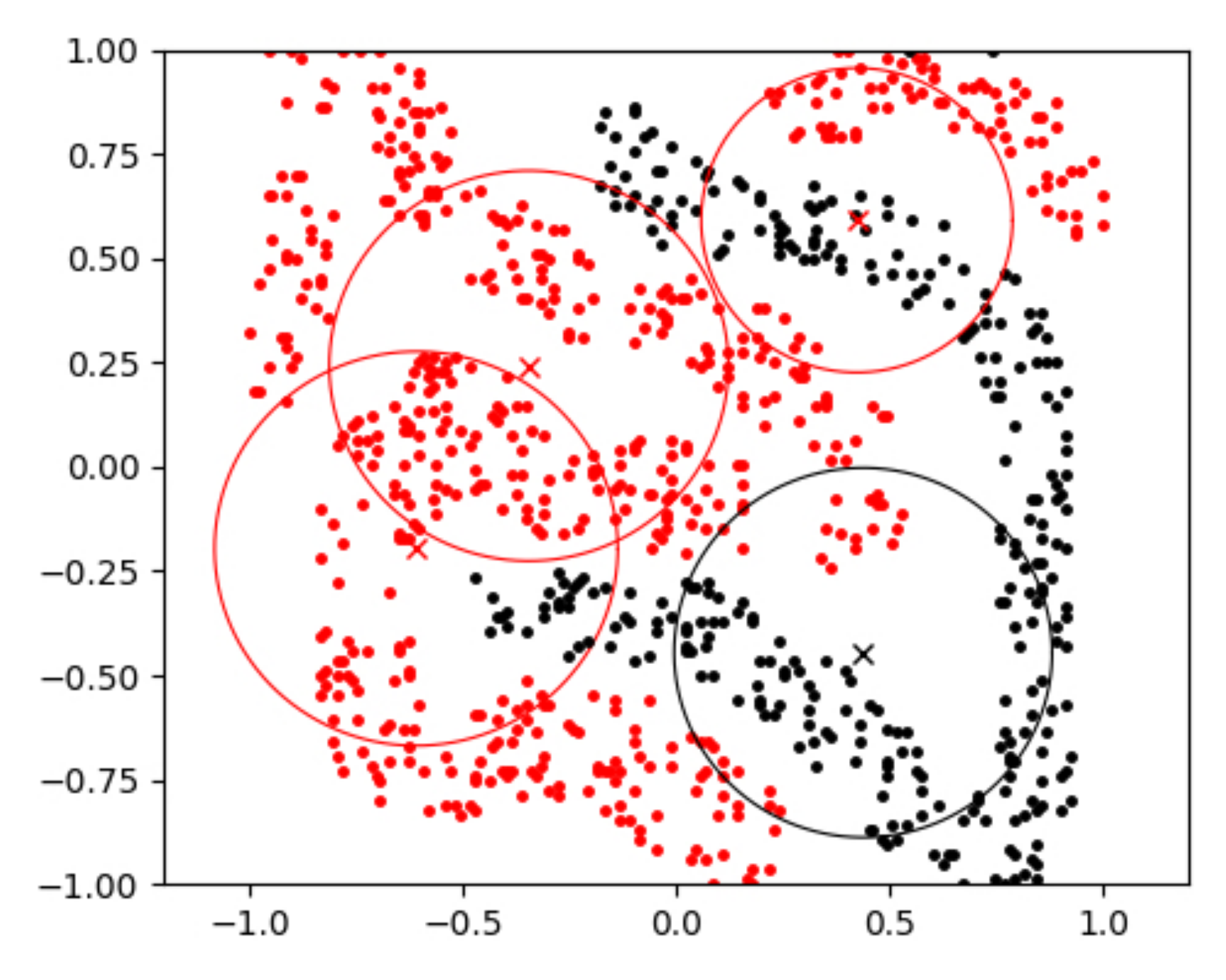}}
		\subfigure[]{\includegraphics[width=0.23\textwidth]{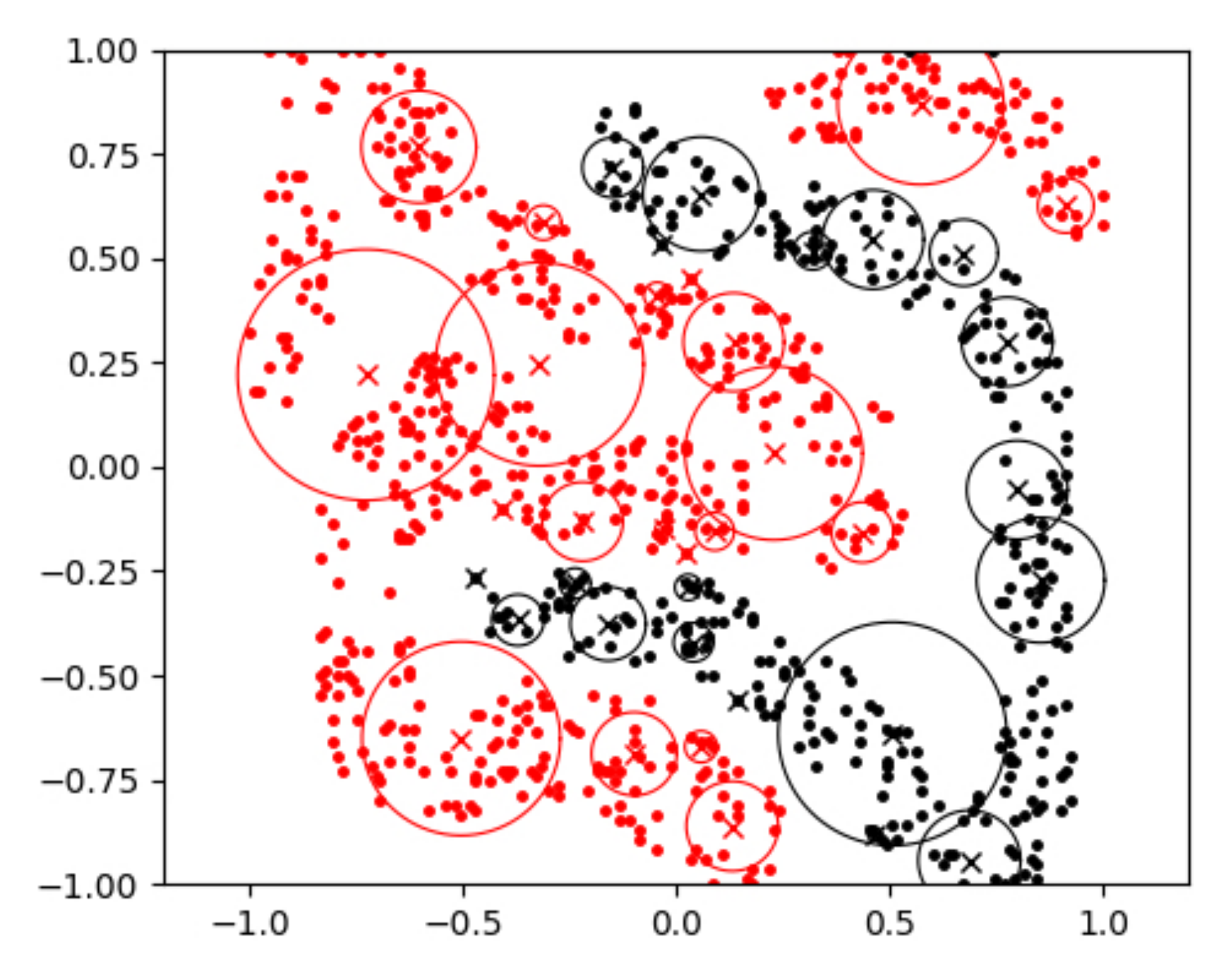}}
		\subfigure[]{\includegraphics[width=0.23\textwidth]{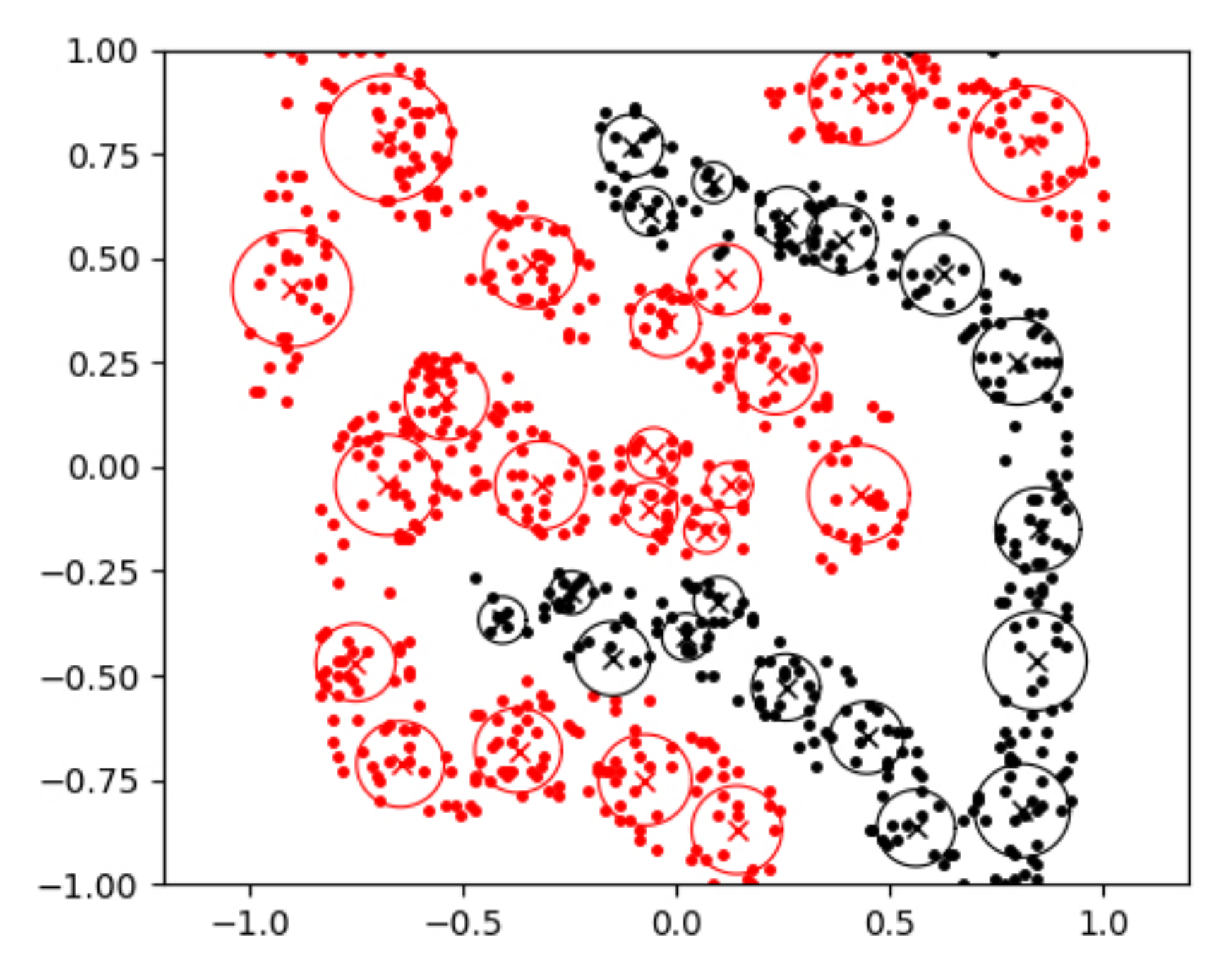}}
		\subfigure[]{\includegraphics[width=0.23\textwidth]{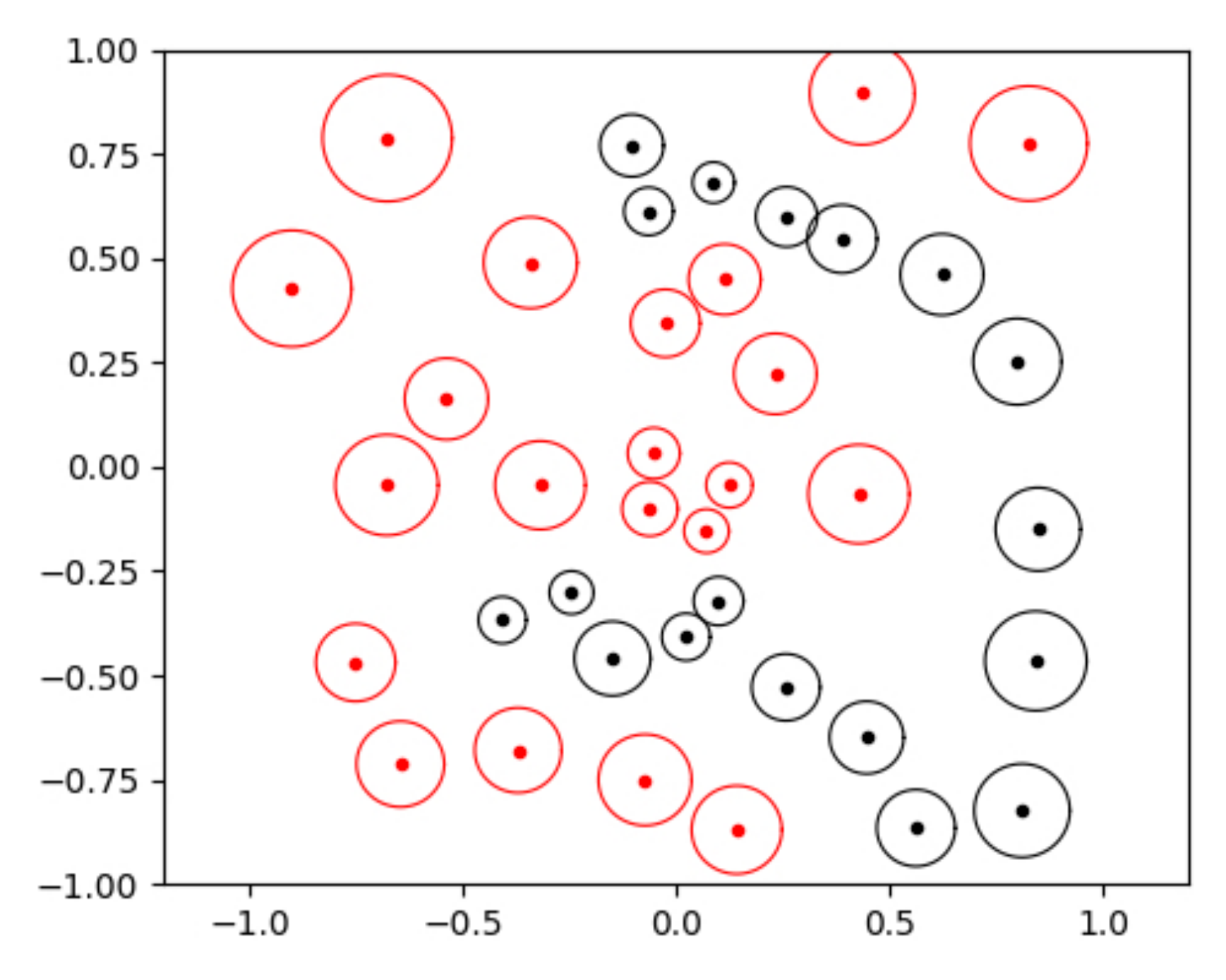}}
		\centering{\caption{The granular-ball splitting generation process of the existing method on the data set fourclass. The colors of the two granular-balls in the figure (corresponding to the two sample point colors) respectively represent the two types of category labels. (a) The initial granular-ball, the whole data set can be seen as a granular-ball to participate in subsequent iterations;(b) Granular-balls generated in the first iteration; (c) Granular-balls generated in the second iteration; (d) Stop splitting results; (e) Results after stopping splitting; (f) Granular-balls extracted.}
			\label{Fig8}}
		\vspace{-1em}
	\end{figure}

	An example of granular-ball generation on the data set fourclass is shown in Fig. \ref{Fig8}. At the beginning of the algorithm, as shown in Fig. \ref{Fig8}(a), the whole data set can be seen as a granular-ball. As the fourclass contains two classes of points, the $k$ in the step 2 in Fig. \ref{fig:GBCProcess} is equal to 2, and two heterogeneous points are randomly selected as the initial centers of two child granular-balls. The experimental results are shown in Fig. \ref{Fig8}(b). However, the granular-balls are too coarse, and the qualities of granular-balls are not high enough, i.e., that their purities do not reach the purity threshold. So, the decision boundary of the granular-balls is inconsistent with that of the data set. As the splitting process progresses, as shown in Fig.\ref{Fig8}(c)-(d), the granular-balls change to be fine, and the purity of each granular-ball becomes to be high until it reaches to the purity threshold or other quality measurement. As shown in Fig. \ref{Fig8}(e), each granular-ball reaches to the purity threshold, and is enough fine. At this time, the decision boundary is very consistent with that of the data set. Fig. \ref{Fig8}(f) shows the extracted granular-balls when the points are removed.
	
	The granular-ball computing has developed granular-ball classifiers~\cite{17xia2019granular}, granular-ball clustering~\cite{16xia2020fast}, granular-ball neighborhood rough set~\cite{8xia2020gbnrs} and granular-ball sampling methods~\cite{37xia2021granular}.
	
	\subsection{Granular-ball k-Nearest Neighbor}
	The $k$NN algorithm has many characteristics such as simple, natural response to multiple classifications, independent of training, can be used for classification and regression at the same time, and easy to parallelize and implement. It is one of the most widely used artificial intelligence algorithms. In $k$NN, the basic principle of finding the $k$ nearest neighbors of a query point is to compute the Euclidean distance from the query point to all data points, and use the values (or labels) of these neighbors to predict or classify the query point. This method is called Full Search Algorithm (FSA). FSA has the following common problems: it needs to optimize the value of $k$; and the optimization of the value of $k$ requires quadratic time complexity, so it is very time-consuming. From the perspective of multi-granularity, the source of the problem of $k$ optimization is due to excessive attention to fine-grained. For this reason, in~\cite{17xia2019granular}, we introduce granular-ball into $k$NN and propose an efficient nearest neighbor algorithm without optimizing $k$ value, granular-ball $k$ nearest neighbors method (GB$k$NN). The basic idea is very easy to implement: base on the granular-ball computing, a single query sample point is a granular-ball with a very small radius, and its predicted label is equal to the nearest granular-ball's label, which is determined by the majority labels in the granular-ball. Therefore, the common feature of GB$k$NN and traditional $k$NN is that the mark of the query point is determined by many points. But the important advantage of GB$k$NN is that there is no need to optimize the parameter $k$, and the label of the query point is determined by adaptively generated nearest neighbor granular-ball with different coarse grains; the second advantage of GB$k$NN is that the number of granular-balls is much smaller than the sample points, and the calculation amount of the nearest granular-ball queried by the query point is much less than $k$NN, resulting in higher efficiency of GB$k$NN than traditional $k$NN; the third advantage is that the decision of traditional $k$NN will be affected by label noise, but GB$k$NN can obtain higher accuracy because of its robustness especially on noisy data sets. These three points are important advantages of GB$k$NN.		
	
	\vspace{-0.5em}
	\subsection{Granular-ball Sampling}
\begin{figure}[!t] 
		\centering
		\subfigure[]{\includegraphics[width=0.23\textwidth]{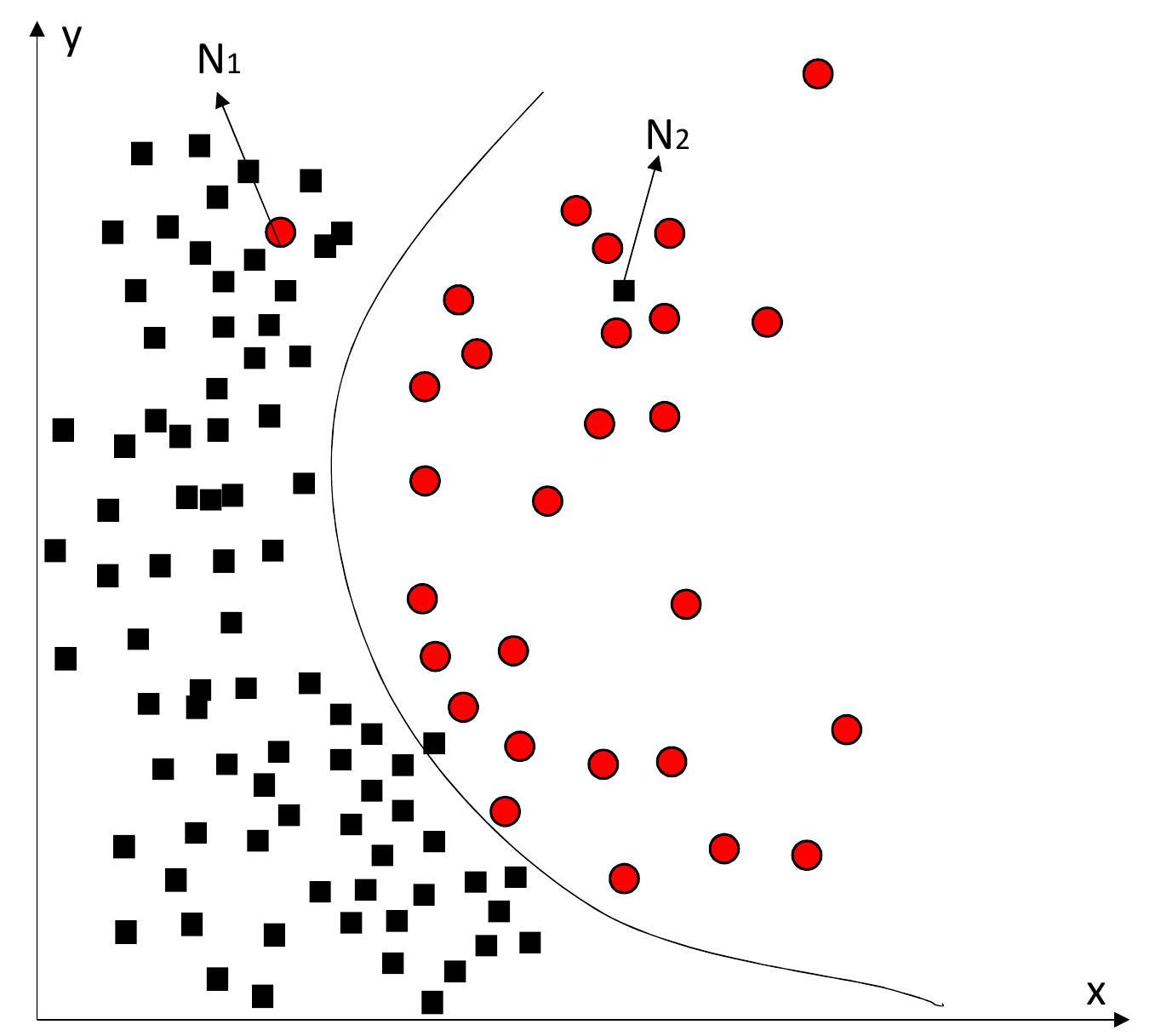}}
		\subfigure[]{\includegraphics[width=0.23\textwidth]{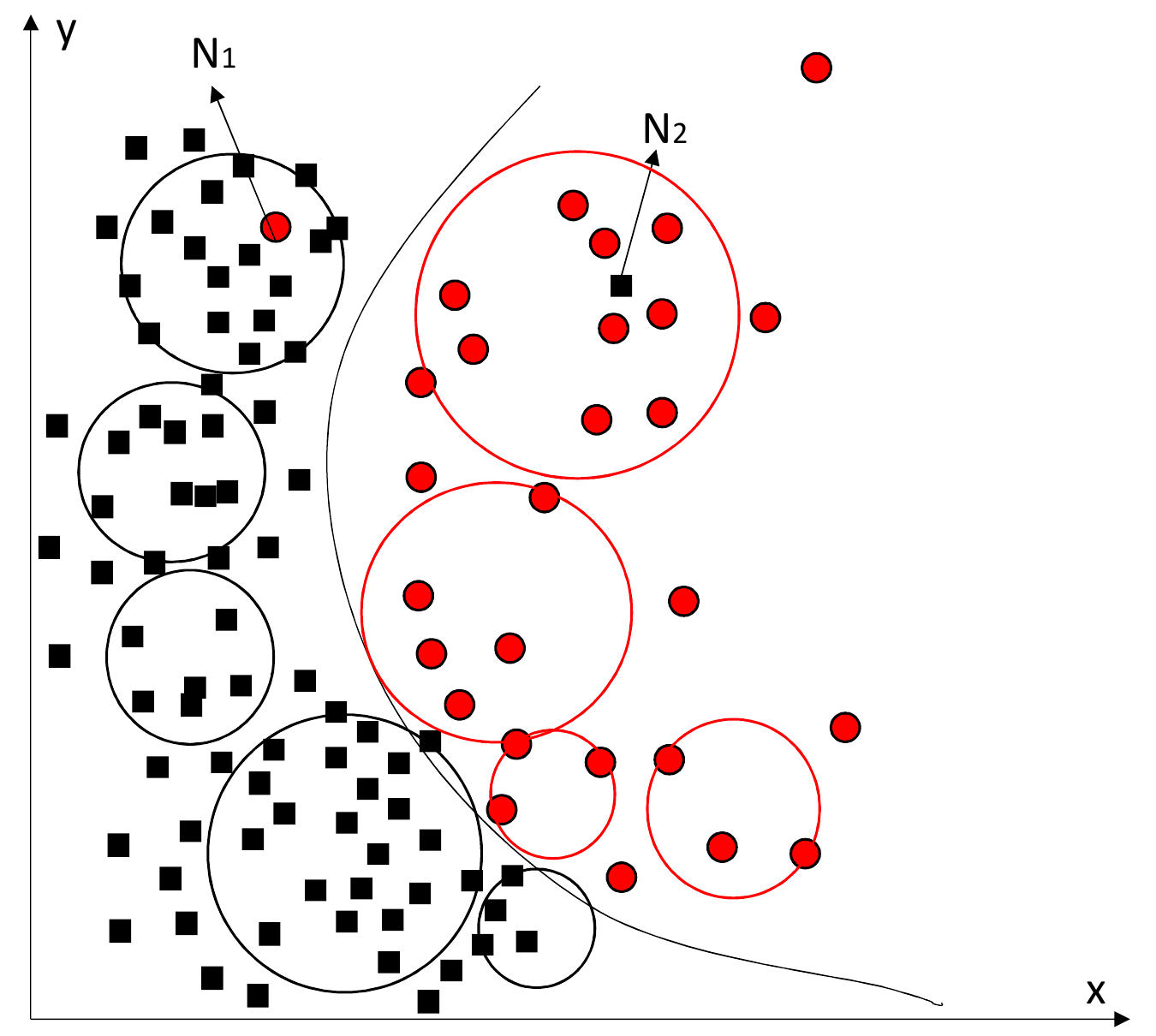}}
		\subfigure[]{\includegraphics[width=0.23\textwidth]{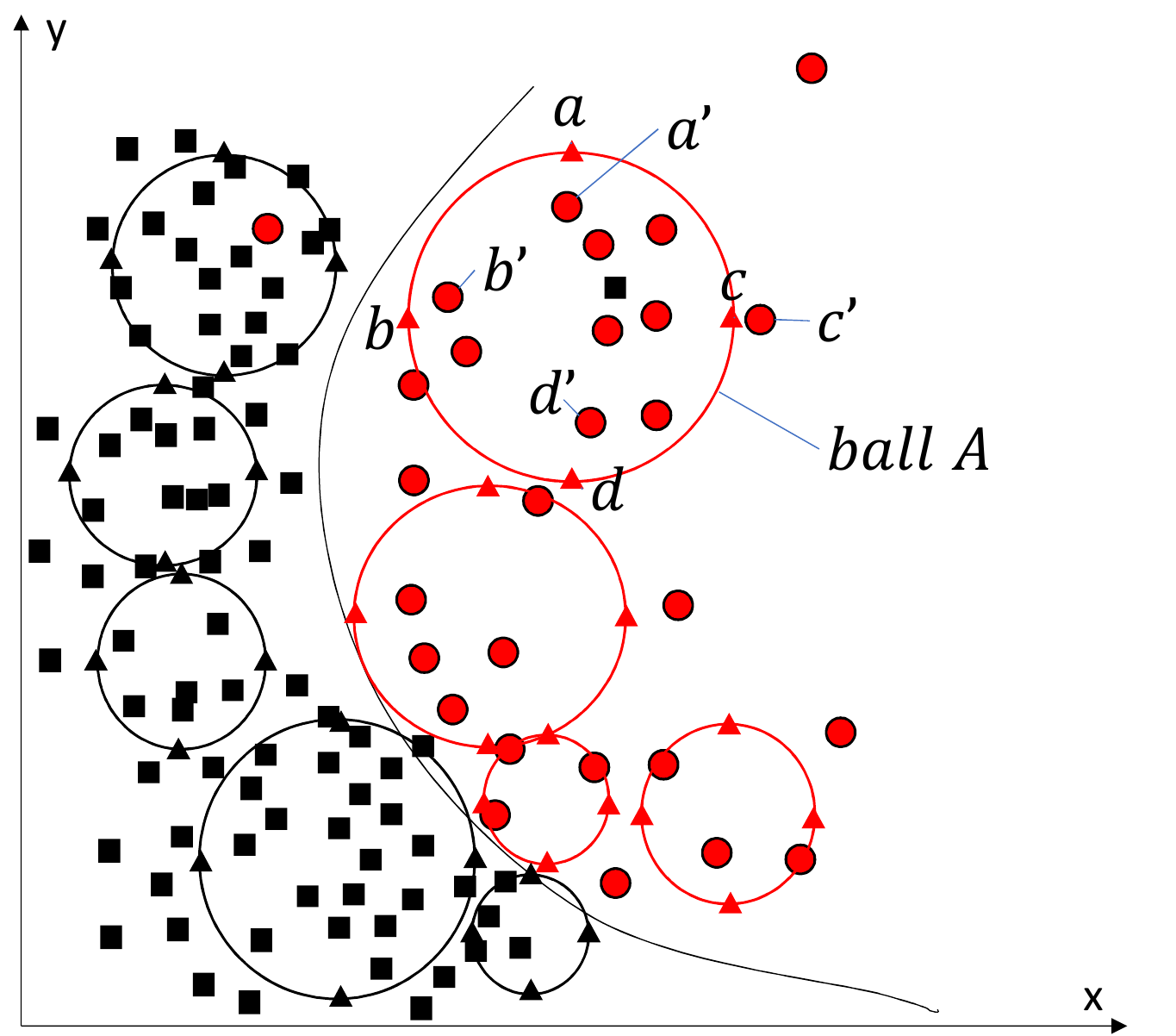}}
		\subfigure[]{\includegraphics[width=0.23\textwidth]{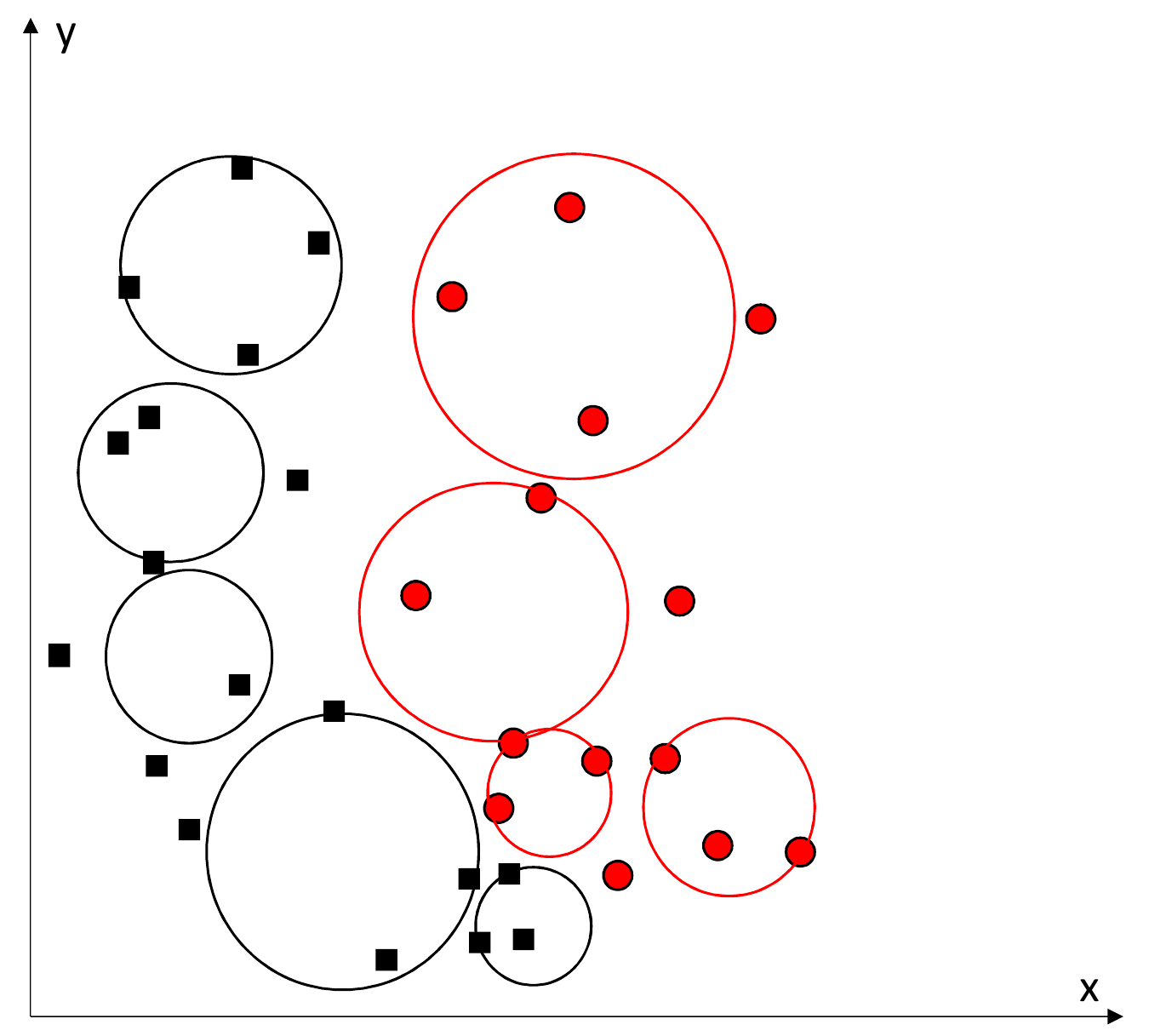}}
		\subfigure[]{\includegraphics[width=0.23\textwidth]{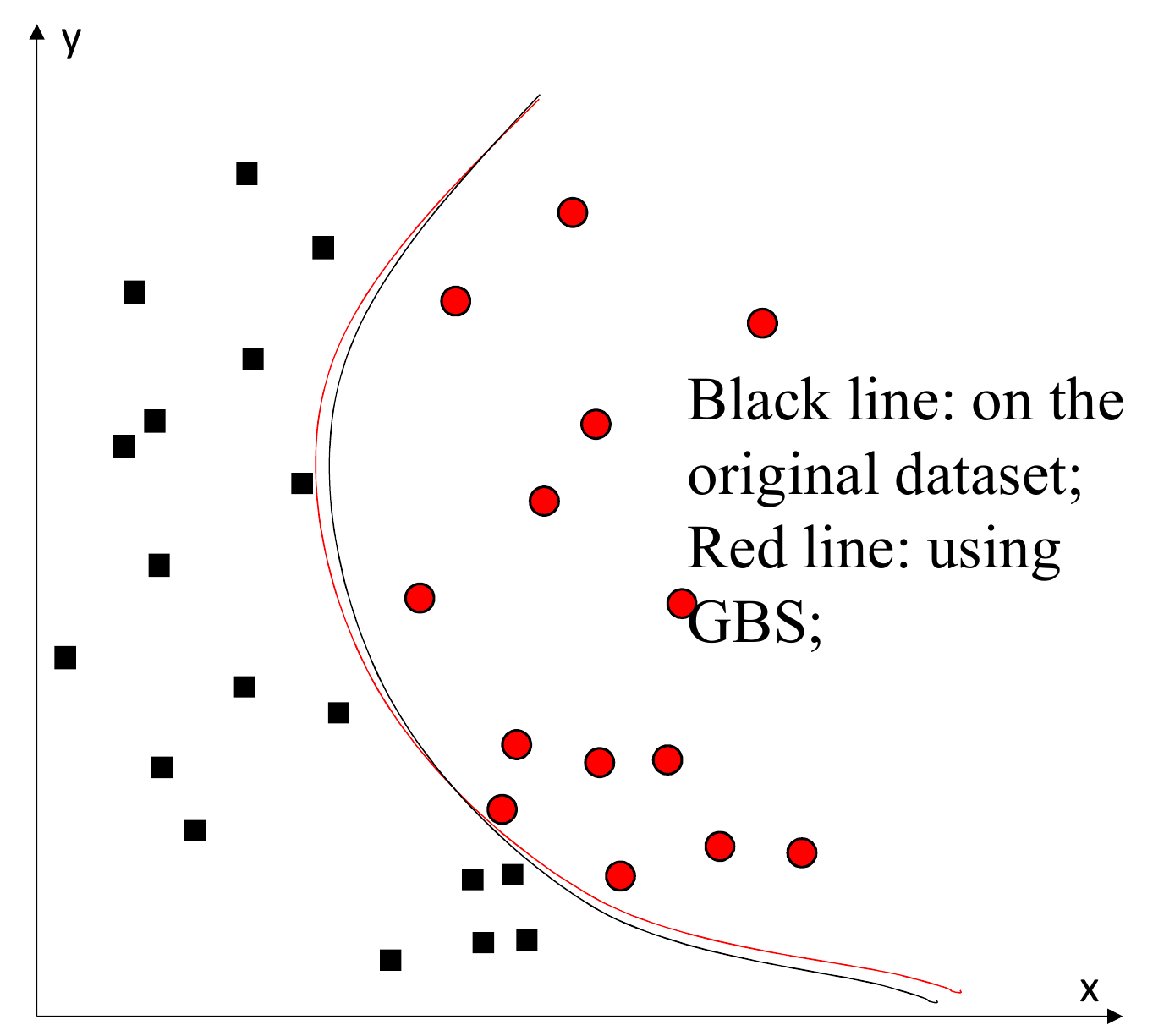}}
		\subfigure[]{\includegraphics[width=0.23\textwidth]{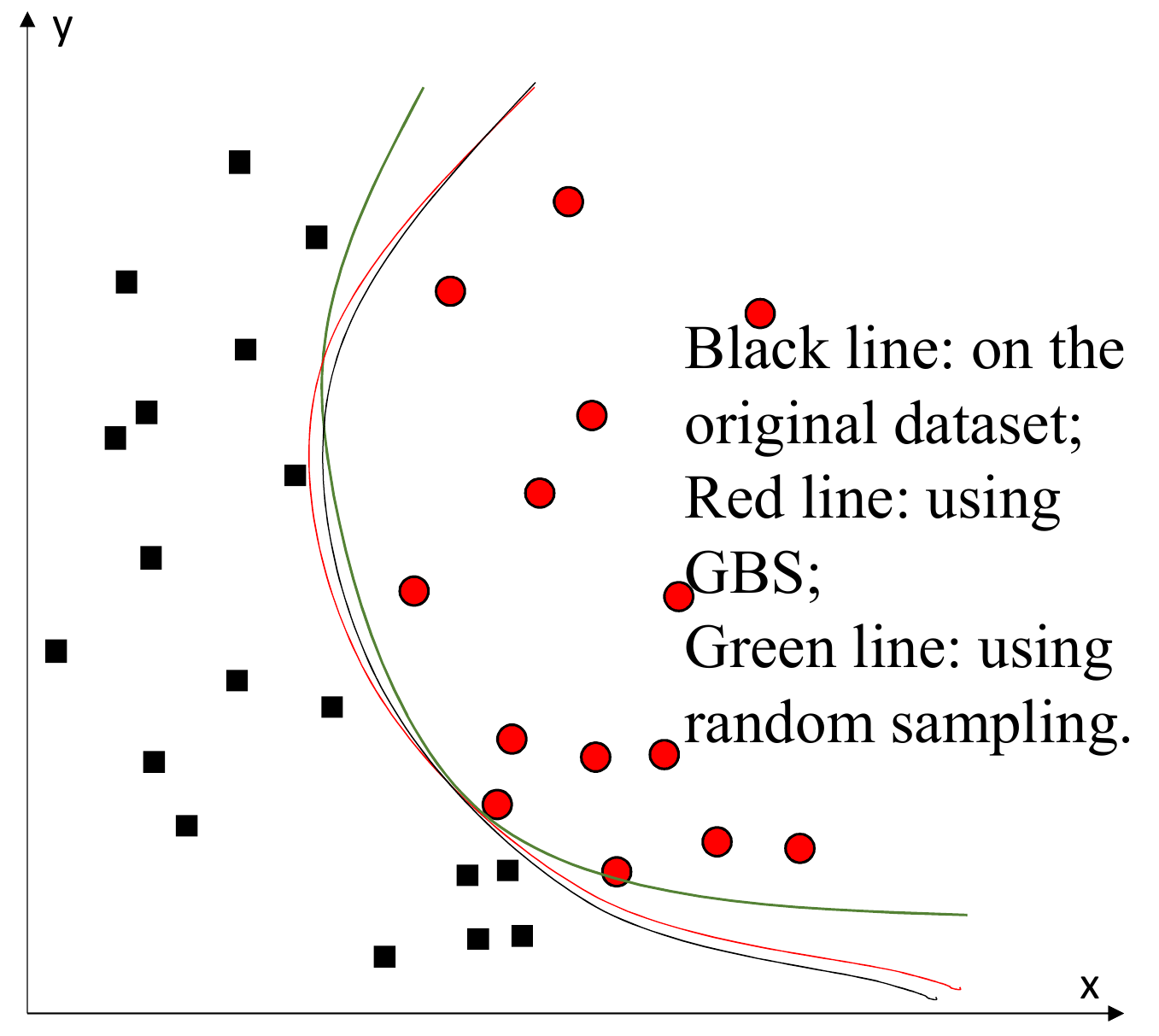}}
		\centering{\caption{~\cite{37xia2021granular} The schematic diagram of GBS. (a) The original data set and its boundaries; (b) The granular-balls generated; (c) The intersection points $a, b, c$ and $d$ in 2 times the dimensions. And points $a', b', c'$ and $d'$ closest to the intersection points in the granular-ball A;(d) use granular-balls to sample the original data; (e) The final data set and its boundaries; (f) The sampling results come from random sampling.}
			\label{Fig4}}
		\vspace{-1em}
	\end{figure}

	The purpose of granular-ball sampling (GBS) is to decrease the size of a dataset in classification by introducing the idea of granular-ball computing. The GBS method uses some adaptively generated balls to cover the data space, and the points near the boundary of each  granular-ball constitute sampled results~\cite{37xia2021granular}. Fig. \ref{Fig4} shows the basic idea of GBS. Fig. \ref{Fig4}(a) shows the original data set and its decision boundary; Fig. \ref{Fig4}(b) shows the original data set after being covered by the granular-balls. In Fig. \ref{Fig4}(c), we find the intersections of the coordinate axis with the granular-ball center as the origin and the granular-ball. In the granular-ball, among the points with the same label as the ball, the points closest to these intersections constitute the sampled result. These points are near the boundaries of the granular-ball. The same label can filter the affection of label noise points. For example, for granular-ball $A$ in Fig. \ref{Fig4}(c), the intersection points of the ball and the coordinate axis with the center of the ball as the origin are $a, b, c$ and $d$, and the points with the same label as the ball closest to these intersections are $a', b', c'$ and $d'$ in $A$ . Therefore, $a', b', c'$, and $d'$ are the sampling results in $A$, and they are also the best points to describe the boundary of the granular-ball $A$. Fig. \ref{Fig4}(d) shows the sampling results of the whole data set. Comparing Fig. \ref{Fig4}(a) and Fig. \ref{Fig4}(e), it can be observed that the boundary curve in Fig. \ref{Fig4}(e) is very consistent with the boundary curve in Fig. \ref{Fig4}(a). At the same time, the samples in Fig. \ref{Fig4}(e) are less and sparser than those in Fig. \ref{Fig4}(a). In contrast, random sampling is very easy to cause the loss of boundary information. Therefore, as shown in Fig. \ref{Fig4}(e) and Fig. \ref{Fig4}(f), the boundary generated by GBS is closer to the boundary of the original data set than the boundary generated by random sampling.
	
	For the intersection point $a$ on the granular-ball, its sampled point $a'$, the closest homogeneous point of $a$, has the same label with the granular-ball. The intersection point can be expressed as a point where the center point vector $c$ moves along the specified coordinate axis by a length $r$. The moving direction of the center point vector includes a positive direction and a negative direction, so one coordinate axis can correspond to two intersection points. Specifically, for a $d$-dimensional data set $D$, the center point vector $c$ of the $i^{th}$ granular-ball generated on $D$ is $ c=\left ( c_{i}^{1}, c_{i}^{2}, \cdots , c_{i}^{j}, \cdots , c_{i}^{d}\right ) $ , the radius $r_{i}$. The two intersection points of the positive and negative directions of the $j^{th}$ coordinate axis can be expressed as
	
	\begin{equation}
		b_j^+=\left ( c_{i}^{1}, c_{i}^{2},\cdots, c_{i}^{j}+r_{i},\cdots, c_{i}^{d}\right ),
	\end{equation}
	\begin{equation}
		b_j^-=\left ( c_{i}^{1}, c_{i}^{2},\cdots, c_{i}^{j}-r_{i},\cdots, c_{i}^{d}\right ).
	\end{equation}
	
	In a data sets, the sampled result $S_k$ for the $k^{}$ granular-ball can be modeled as follows:
	\begin{scriptsize}
		\begin{equation} \label{equ:gbs}
			\begin{aligned}
				S_k= \left\{min\left \{ dis\left ( x_i, b_1^+ \right ), x_i\in GB_k,label(x_i)==label(GB_k) \right \},
				\right.
				\\
				\phantom{=\;\;}\left. min\left \{ dis\left ( x_i, b_1^- \right ), x_i\in GB_k,label(x_i)==label(GB_k)  \right \},
				\right.
				\\
				\phantom{=\;\;}\left.\cdots \qquad \qquad \qquad \qquad \qquad \qquad \qquad  \qquad
				\right.
				\\
				\phantom{=\;\;}\left. min\left \{ dis\left ( x_i, b_d^+ \right ), x_i\in GB_k,label(x_i)==label(GB_k)  \right \},
				\right.
				\\
				\phantom{=\;\;}\left. min\left \{ dis\left ( x_i, b_d^+ \right ), x_i\in GB_k,label(x_i)==label(GB_k)  \right \} \right\},
			\end{aligned}   
		\end{equation}
	\end{scriptsize}
	where $S_k$ is sampled result for the granular-ball $GB_k$, and the label of $S_k$ is consistent with that of $GB_k$. For a unbalanced data set, the minority class of points are remained, and the sampling process is performed only on the majority class of points. So, in the Equ. (\ref{equ:gbs}), $S_k\in GB_k$ needs to be changed to $S_k\in majGB_k$, which represents the majority granular-balls.

	In the classification with label noise, GBS can both reduce a data set and improve the its data quality. Besides, GBS is also effective for undersampling of unbalanced classification. In addition, the time complexity of GBS is O($n$), so it can speed up most classifiers~\cite{38zhou2009novel}.
	
	\section{Granular-ball Covering Model}
	At present, the granular-ball covering of granular-ball computing lacks of mathematical model. In this section, we will establish the basic model of granular-ball covering, and the specific description is as following: given a data set $D=\{x_i (i=1,\cdots ,n)\}$, where $n$ is the number of samples on $D$. Granular-balls $GB_1, GB_2, \cdots, GB_m$ are used to cover and represent the data set $D$. The original goal of the optimization problem of the granular-ball generation method is expressed as $OO_{bj}$, and the main factors of measuring the coverage are: 1. Coverage degree, when other factors remain unchanged, the higher the coverage, the less sample information is lost and the more accurate the characterization. Suppose the number of samples in the $j^{th}$ granular-ball $GB_j$ is expressed as $|GB_j|$, then its coverage degree can be expressed as ${\textstyle \sum_{j=1}^{m}( | GB_j | ) } /n$. 2. When other factors remain unchanged, the number of the granular-balls is related to the size of granular-ball. The minimization of this factor is to make the granular-balls as coarser as possible. The fewer the number of granular-balls, the coarser the granular-balls and the more coarse-granularity characteristics: the more efficient the granular-ball calculation is, the better the robustness is. 3. In addition, under different problems, in order to correspond to the relevant optimization goal $OO_{bj}$, the quality of the granular-ball $GB_j$ needs to be higher than a given purity threshold $T$ of the given evaluation method. This factor is also related to the lower limit of the size of the granular-balls, so that the granular-balls must be ``fine" enough to accurately describe the problem. The threshold can be obtained in a given way, or in a lattice search, or in an adaptive way, which is pursued by us. Taking the reciprocal of the granular-ball covering to optimize its minimum value and the optimization goal of the granular-balls can be expressed as
	
	\begin{equation}
		\begin{aligned}
			&Min\ \lambda _1*n/\sum_{j=1}^{m} (GB_j)+\lambda _2*m, \\ 				
			&s.t.\ quality(GB_j)\ge T,
			\label{e0}
		\end{aligned}
	\end{equation}	
	where $\lambda _1$ and $\lambda _2$ are the corresponding weight coefficients and $m < n$.
	
\begin{figure}[!t]
		\centering
		\subfigure[]{\includegraphics[width=0.23\textwidth]{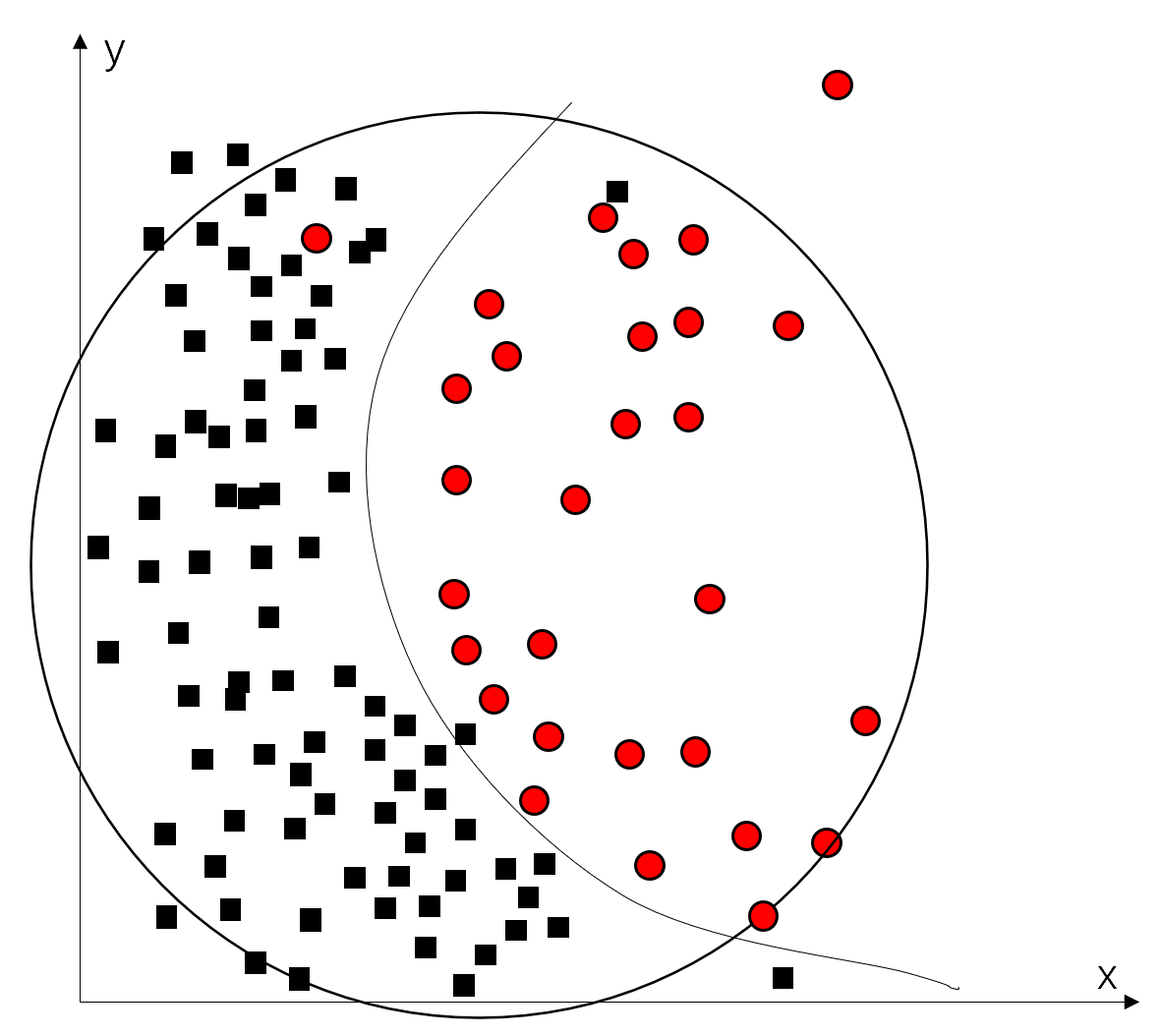}}
		\subfigure[]{\includegraphics[width=0.23\textwidth]{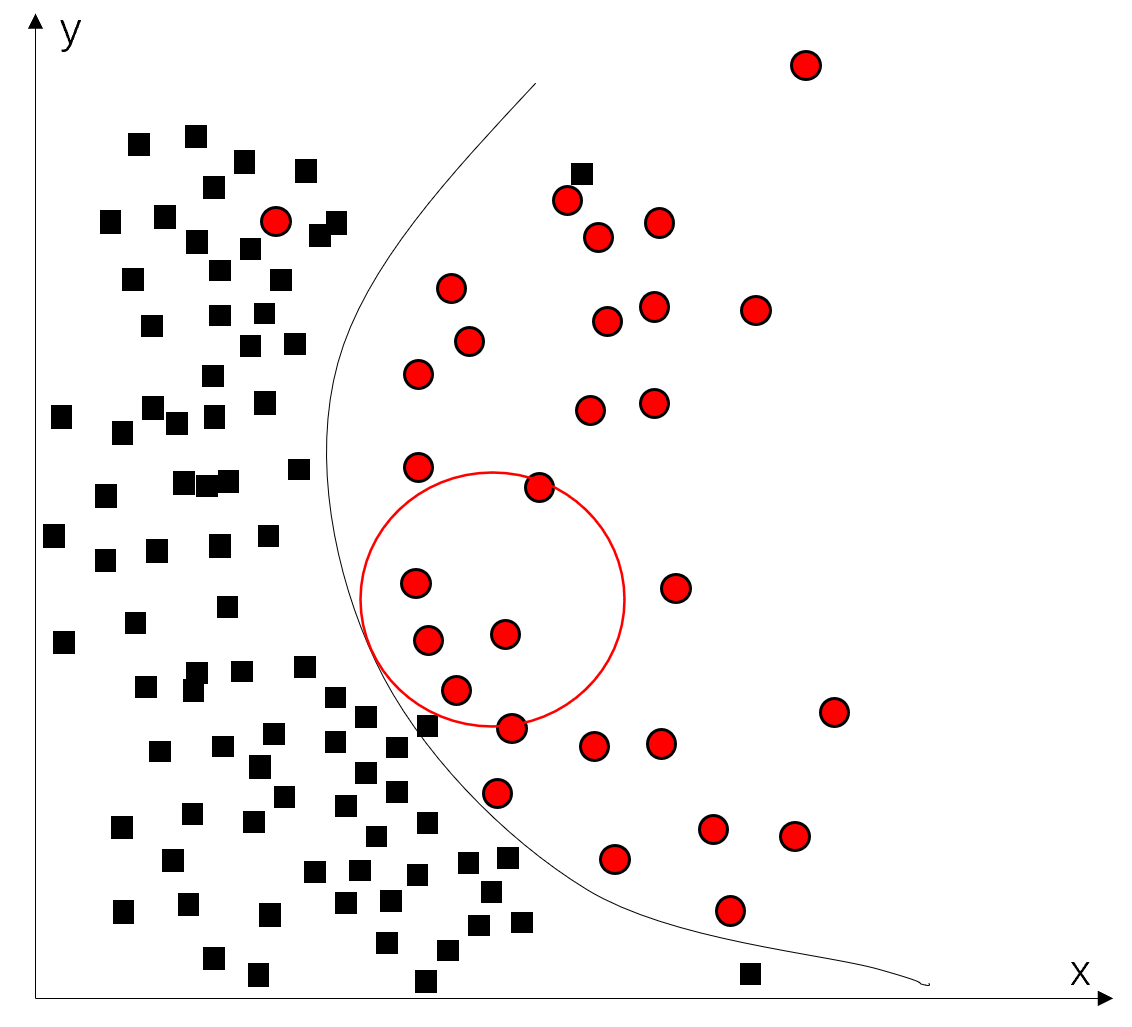}}
		\centering{\caption{Invalid coverage of sample space by granular-balls. (a) Granular-ball coverage results without considering the quality of the granular-ball; (b) Granular-ball coverage results without considering the rate of the coverage of the granular-balls.}
			\label{Figeight}}
		\vspace{-1em}
	\end{figure}

	The definition of granular-ball’s quality is different according to the environment, but it can basically be defined as a sample label with a certain approximate (or equivalence) relation. For example, in the classification problem, we often use the nearest neighbor to describe this equivalence relation. These factors are indispensable. It is obviously unreasonable to rely only on factor 1, coverage degree, or factor 2, the number of the granular-balls. For example, in the extreme case shown in Fig. \ref{Figeight}(a), only one granular-ball is used to represent it. At this time, the quality of the granular-ball is poor, and one granular-ball cannot describe the distribution of a data set (i.e., data boundary). If the factor 2 is not considered, as shown in Fig. \ref{Figeight}(b), the granular-balls can only cover a small part of the data set, and it is also impossible to describe the data set. If do not consider the factor 3 ``the number of granular-balls" (that is, the size of the granular-balls), and only consider the quality and coverage degree, the granular-balls can be divided into the finest granular-ball, i. e., a granular-ball contains only one sample point. When coarse-granularity does not make any sense. Therefore, none of the above factors are indispensable. On the whole, when factor 1 guarantees a certain coverage, the factors 2 and 3 is to obtain granular-balls with appropriate granular-ball size; when the factors 1 and 2 remain unchanged, the smaller the threshold of factor 3, the easier the quality of the granular-balls will be satisfied (i.e. the coarser the granular-balls, the fewer the number of the granular-balls, and the more efficient computational performance can be obtained). The control of the threshold in factor 3 exhibits the ability of scalability of the granular-ball generation. In fact, the existing granular-ball generation method, as shown in Fig. \ref{fig:GBCProcess}, provides a heuristic optimization strategy.
	
	\section{An Acceleration Granular-ball Generation Method}
	
	\subsection{Motivation}
	The existing granular-ball generation method uses the $k$-means algorithm to split the granular-ball; so, the granular-ball generation is not efficient than $k$-means, which can generate a stable splitting results in each iteration of granular-ball generation. However, the stability in the intermediate process is not needed in the process; what is required is only to generate the granular-balls fulfilling Equ. \ref{e0}, such as that the lower bound should be ensured. 
	
	\vspace{-0.5em}
	\subsection{The Process of the Acceleration Granular-ball Generation Method}

\begin{figure}[htbp!]
		\centering
		\fbox{\includegraphics[scale=0.28]{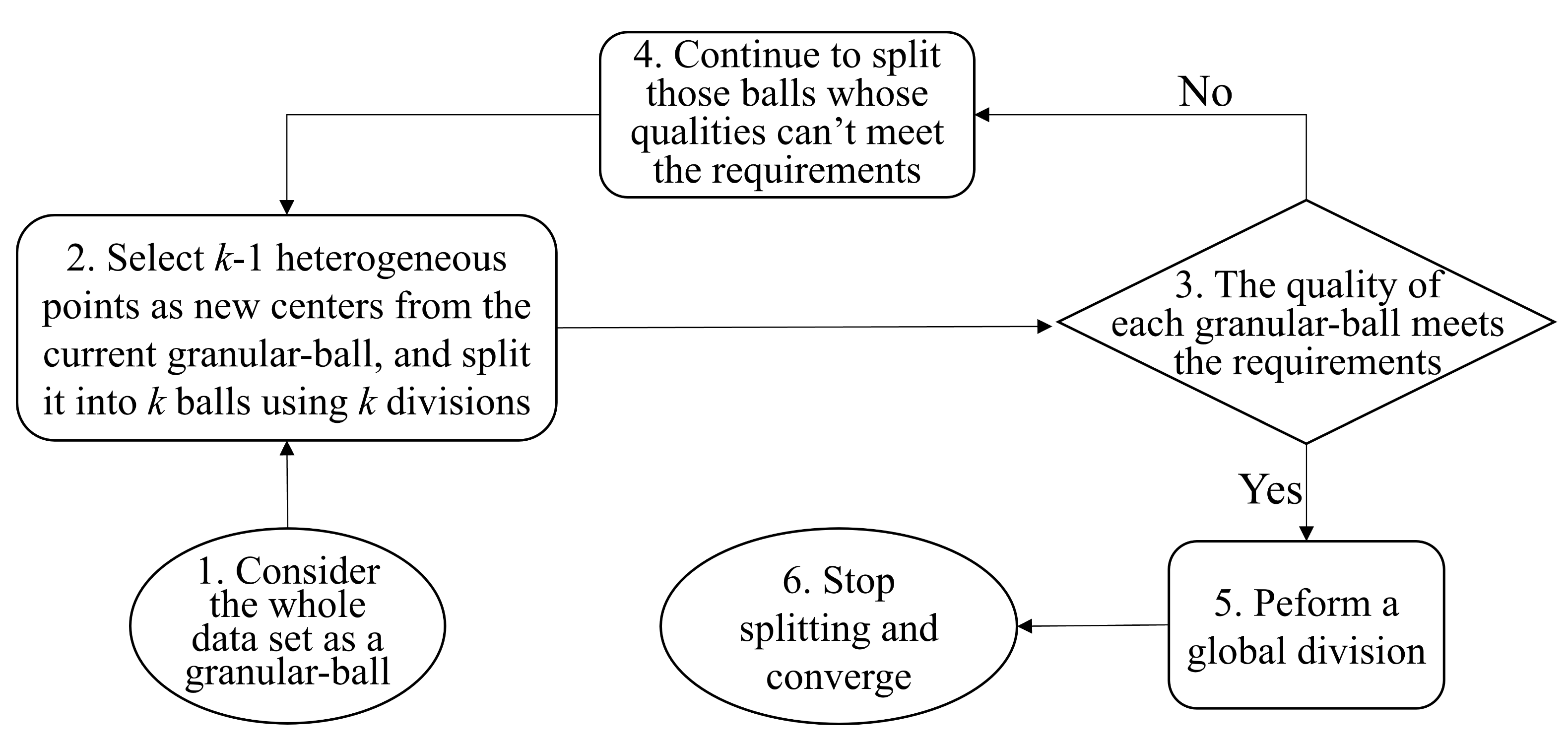}}
		\caption{Process of the acceleration granular-ball generation in granular-ball computing.}
		\label{fig:GBCProcess+}	
	\end{figure}

	As the stability in the intermediate process is not needed, as shown in the step 2 in Fig. \ref{fig:GBCProcess+}, we use one division, i.e., one iteration process in the $k$-means, to split a granular-ball in stead of a whole $k$-means algorithm. Besides, different from the existing method shown in Fig. \ref{fig:GBCProcess}, a global division is added in the end to improve the whole distribution of the final granular-balls. In the global division, a division is performed based on all the division points. The specific process is as Fig. \ref{fig:GBCProcess+}. In order to describe the process of the acceleration granular-ball generation more clearly, we give the definition of the father ball and the child ball at first.
	
	\newtheorem{definition}{\textbf{Definitions}}
	\begin{definition}
		Given granular-balls $A$ and $A_i(i=1,2,\dots,k)$, suppose that $\bigcup_{i=1}^{k}{A_i}=A$ and $\bigcap_{i=1}^{k}{A_i}=\emptyset$. Then $A$ and $A_i$ are the parent ball and the child ball, respectively.
	\end{definition}

	A $k$-means algorithm consists of $t$ iterations, where $t$ denotes the iteration times, and each iteration is a division, in which all points are divided into $k$ clusters according to their distances to the $k$ center points. In the step 2 in Fig. \ref{fig:GBCProcess+}, the $k$-means used for splitting a granular-ball is replaced with $k$ division, where $k$ denotes the number of classes in a granular-ball. So, the computation cost is decreased directly. Besides, as shown in the step 2, taking splitting granular-ball $A$ as an example, the center point of the granular-ball $A$, denoted as $a$, is remained as the center point of a certain child ball of $A$, so only $k$-1 points are selected as centers of those new $k-1$ child balls of $A$. The sample points in all $k$ child balls do not need to calculate the distance from the original center point a because they have been calculated before. Consequently, the computation cost is further decreased.
	
	\begin{figure}[htbp!]
		\centering
		\subfigure[]{\includegraphics[width=0.23\textwidth]{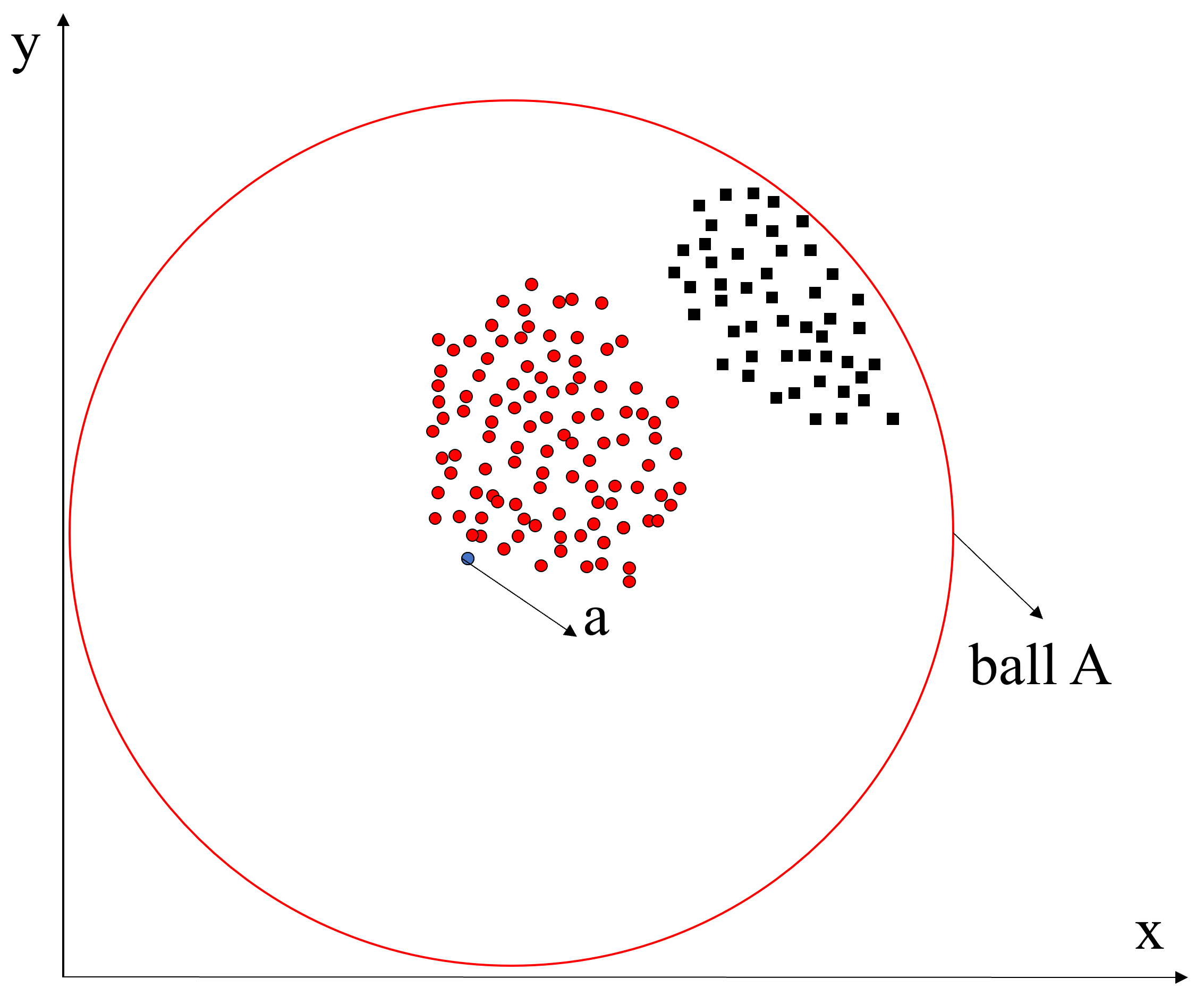}}
		\subfigure[]{\includegraphics[width=0.23\textwidth]{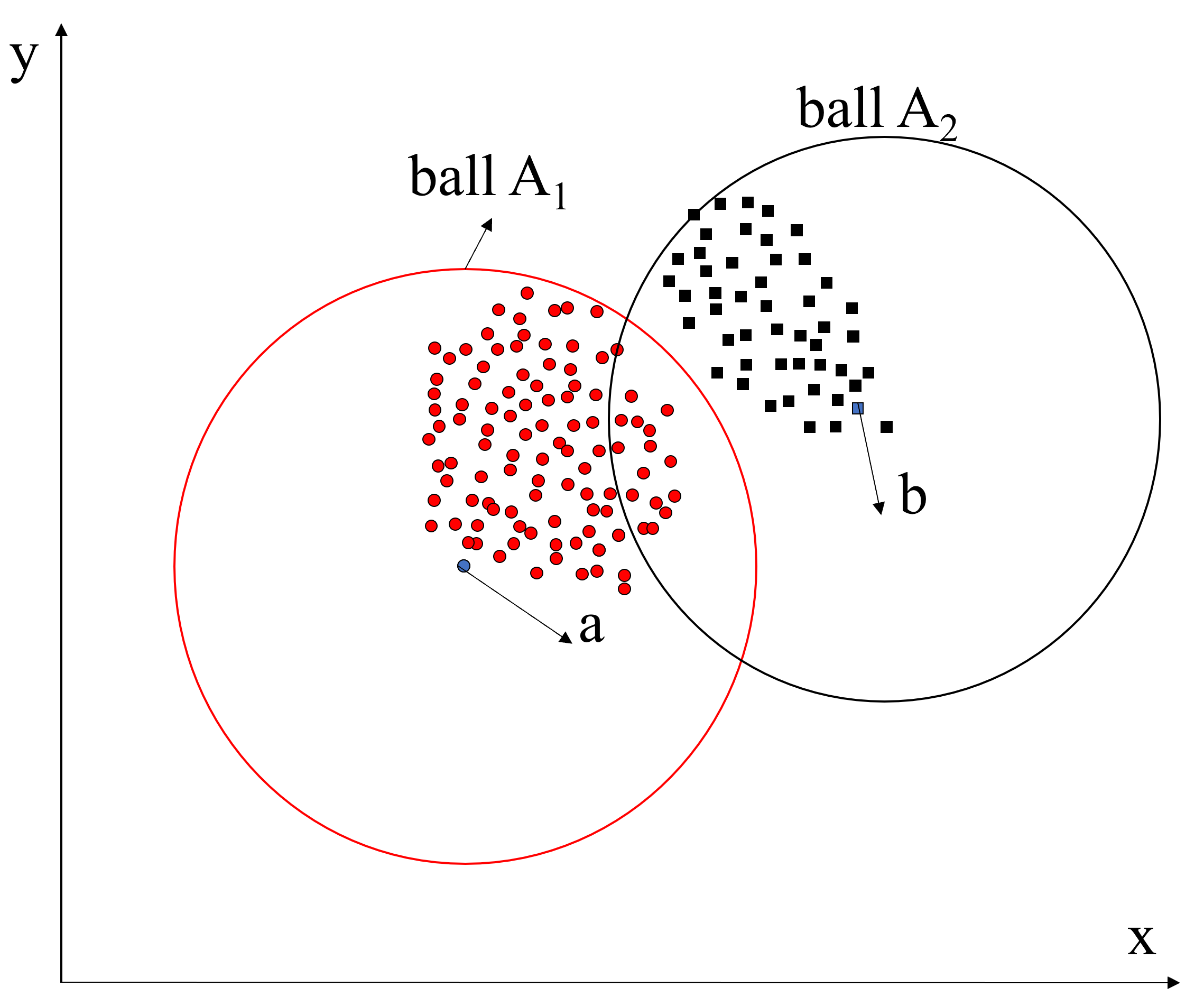}}
		\centering{\caption{Granular-ball splitting using the acceleration granular-ball generation method. (a) granular-ball $A$ with center as $a$; (b) $A$ is split into two child balls $A_{1}$ and $A_{2}$ whose centers are $a$ and $b$ respectively.}
			\label{Fig5}}
		\vspace{-1em}
	\end{figure}
	
	As shown in Fig. \ref{Fig5}, a granular-ball $A$ with center as $a$ is split into two child balls $A_{1}$ and $A_{2}$ whose centers are $a$ and $b$ respectively. The radius of a granular-ball is represented by the furthest distance from the data points in it to its center in order to cover all the data points in the ball. The center $a$ of the ball $A$ in Fig. \ref{Fig5}(a) is remained as the center of child ball $A_1$ in Fig. \ref{Fig5}(b). In this split process of the granular-ball, the distance from all data points in the ball $A$ to the center $a$ does not need to be computed again, and only the distance from all the data points to the center $b$ of child ball $A_2$ in Fig. \ref{Fig5}(b) is computed. Finally, all data points are divided into the two granular-balls based on above distances to the two centers $a$ and $b$.
	
\begin{figure}[!t]
		\centering
		\subfigure[]{\includegraphics[width=0.23\textwidth]{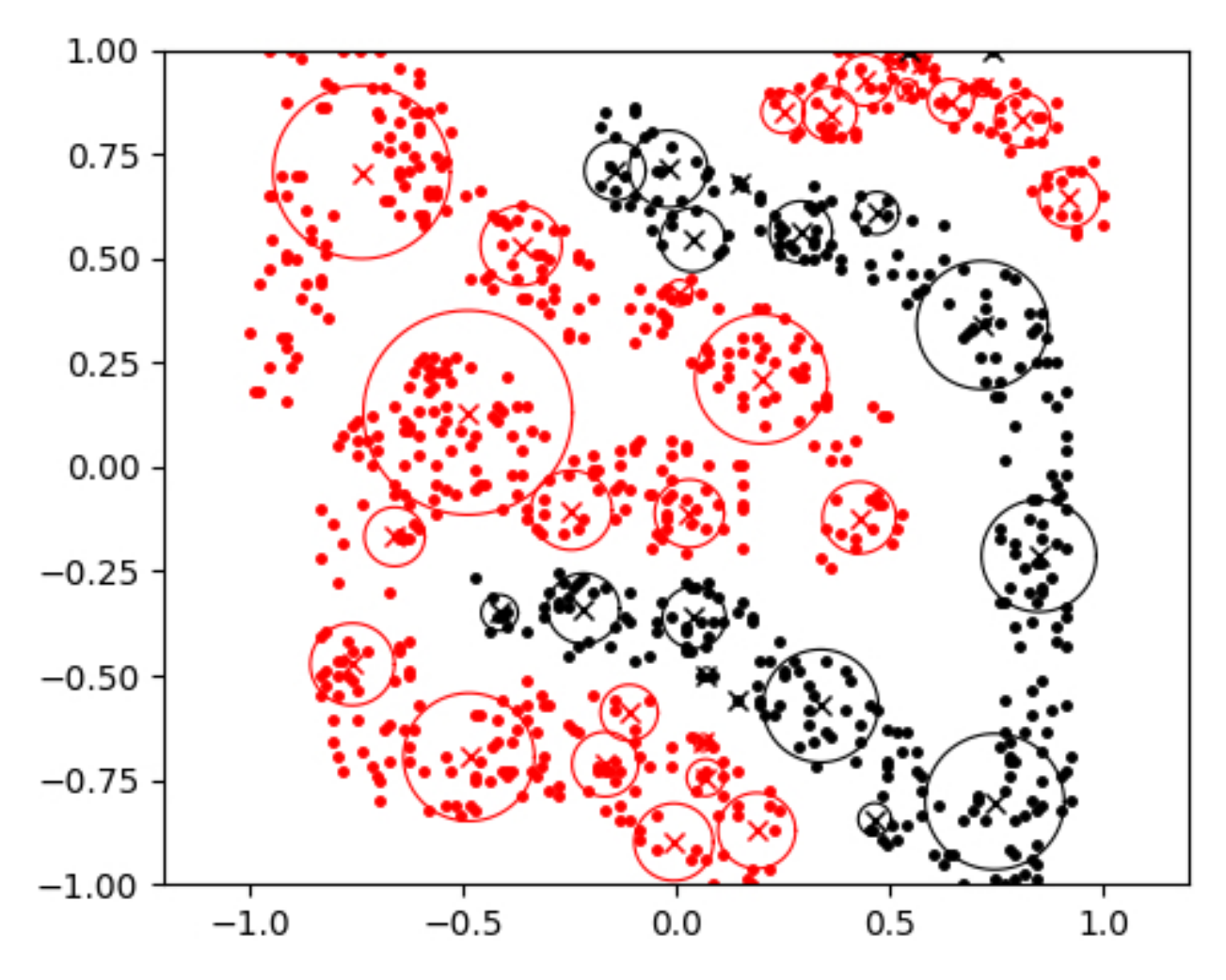}}
		\subfigure[]{\includegraphics[width=0.23\textwidth]{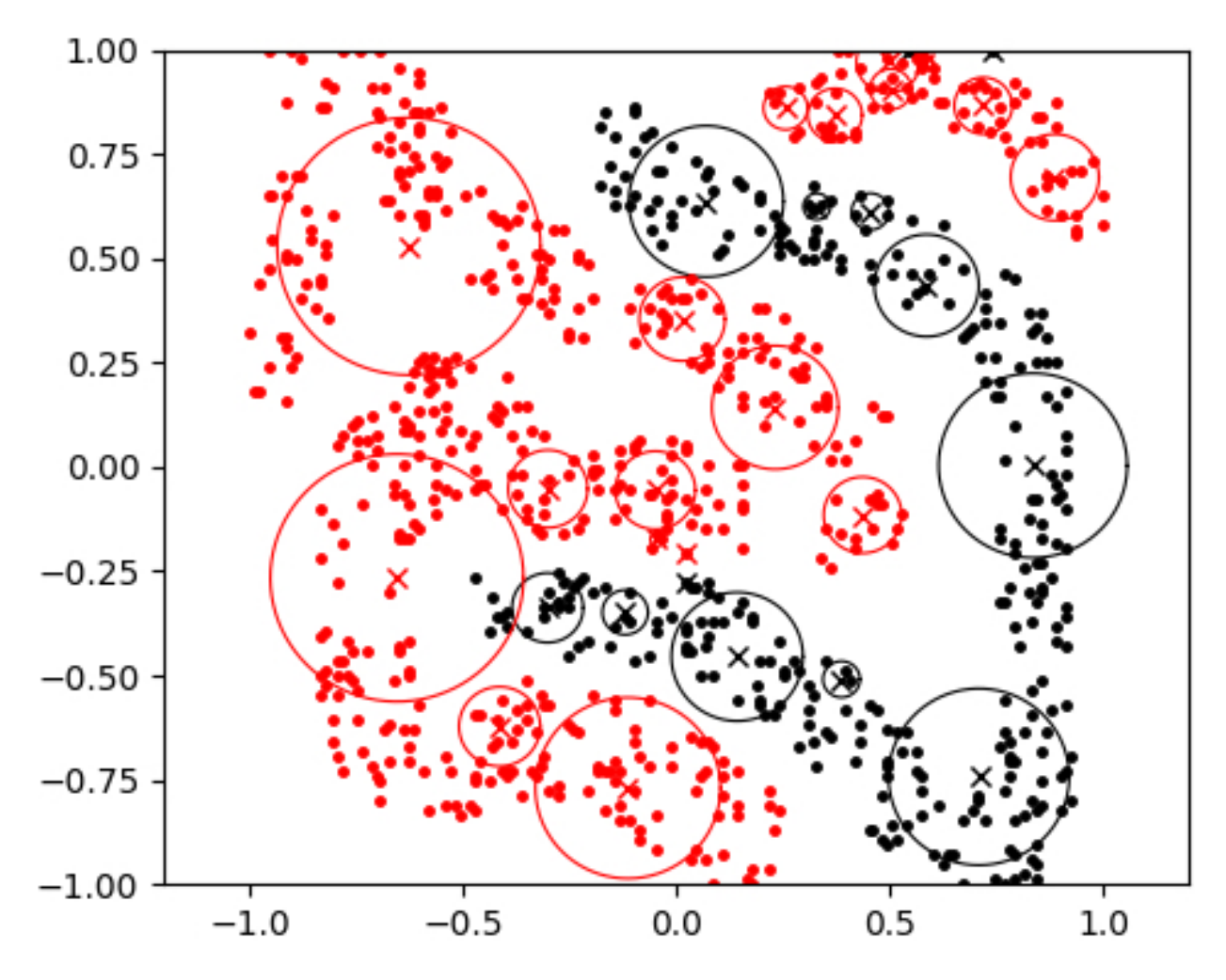}}
		\subfigure[]{\includegraphics[width=0.23\textwidth]{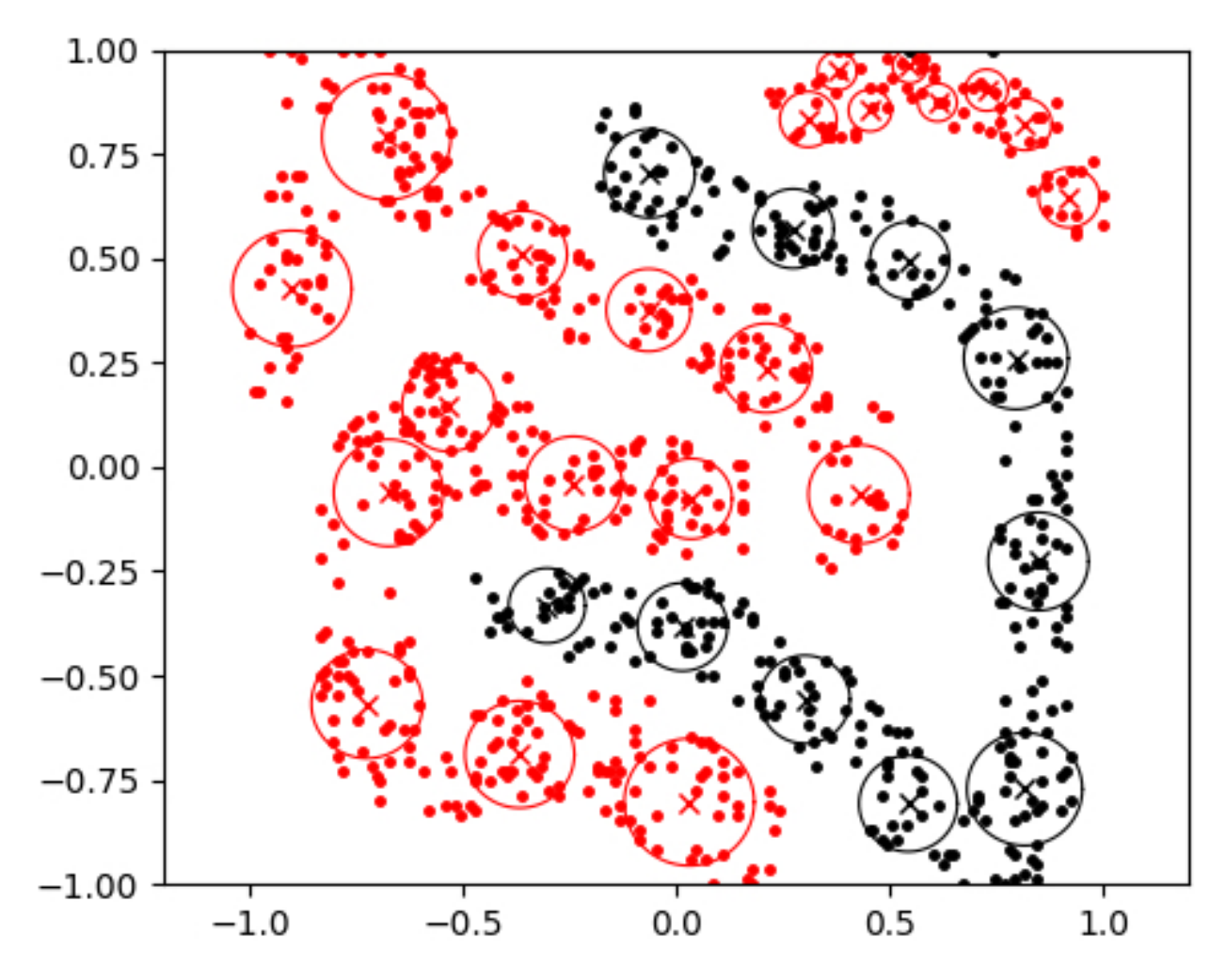}}
		\subfigure[]{\includegraphics[width=0.23\textwidth]{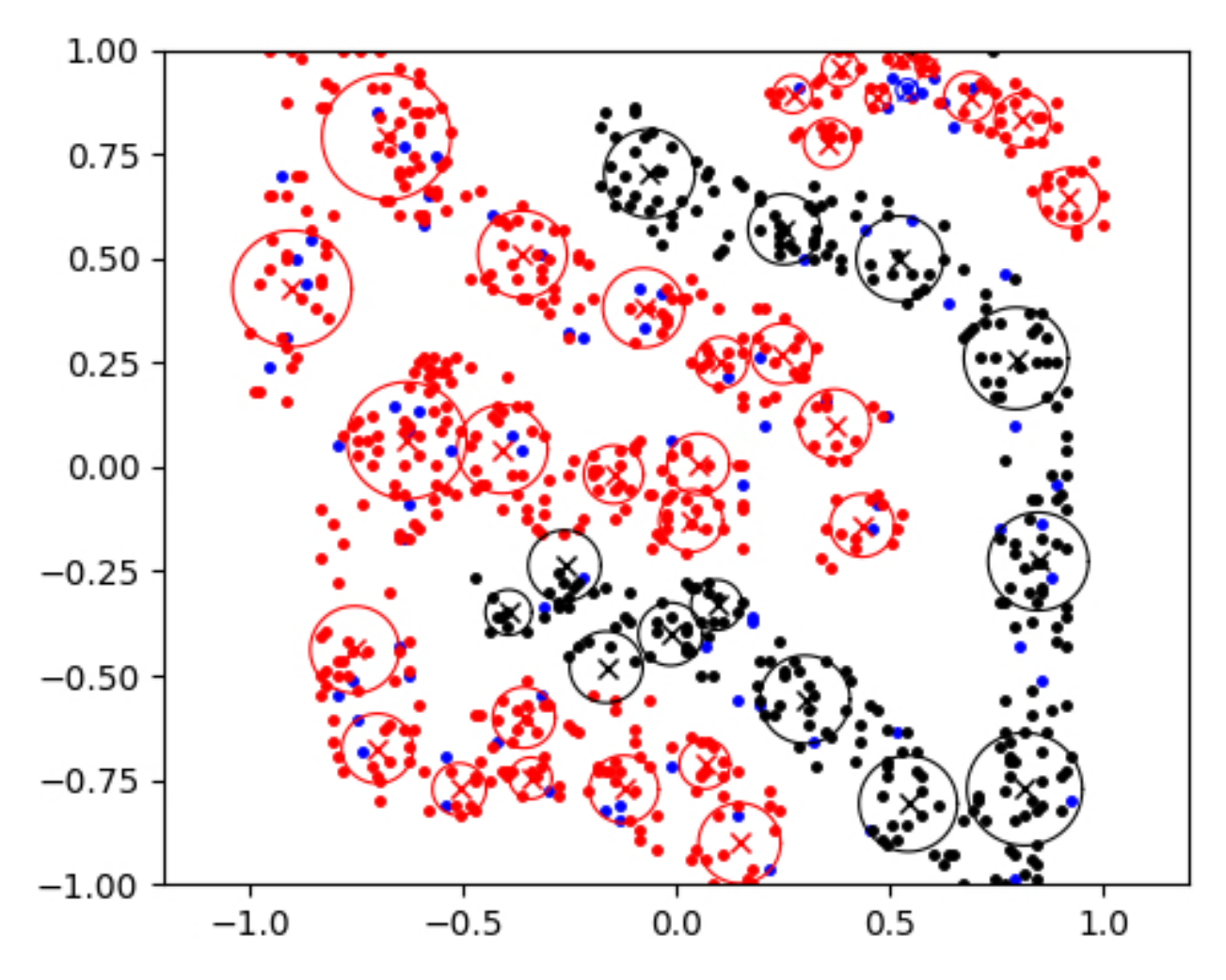}}
		\centering{\caption{Comparison of the distribution of granular-balls before and after global division. (a) The result of granular-ball generation using $k$-means; (b) The result of the acceleration granular-ball generation method without using the global division; (c) The distribution of granular-balls using the global division; (d) The experimental result using the proposed acceleration method in a noisy data set where the noise points are colored with blue.}
			\label{Fig6}}
		\vspace{-1em}
	\end{figure}

	It is worth noting that, in the granular-ball splitting process of the acceleration method, the center of a granular-ball is its division point instead of that computed using Equ. (\ref{equ:center}). As shown in Fig. \ref{Fig5}(a), the center of the granular-ball $A$ is the division point $a$ instead of the center of those data points in $A$, which is computed using Equ. (\ref{equ:center}). However, as shown in the step 7 in Fig. \ref{fig:GBCProcess+}, the global division, i.e., a division on all division points, is performed so that a division point is close to the center of the corresponding granular-ball. For example, Fig. \ref{Fig6} shows the comparison results between the conventional granular-ball generation method and our proposed acceleration method. Fig. \ref{Fig6}(a) shows the granular-ball generation result using the conventional method. Fig. \ref{Fig6}(b) shows experimental results using the proposed acceleration method before the global division is performed. It can be seen from Fig. \ref{Fig6}(c) that, after the global division is performed, in comparison with those in Fig. \ref{Fig6}(b), the division center of a granular-ball, i.e., its division point changes to be closer to its true center computed using Equ. (\ref{equ:center}), so the points in granular-ball becomes to be more tightly and uniformly distributed and the decision boundary is clearer. In addition, Fig. \ref{Fig6}(d) shows the experimental results when some label noise points are added. It an be observed from Fig. \ref{Fig6}(d) that, the granular-balls generated using the acceleration method can fit the division boundary well in the noisy data because of the robustness of granular-ball computing. The algorithm design of the granular-ball generating acceleration method is as Algorithm \ref{alg1}.
	
	\begin{algorithm}[htbp!]
		\caption{Granular-ball generation acceleration method}
		\label{alg1}
		\textbf{Input}: Data set $D$, the purity threshold $T$\\
		\textbf{Output}: The granular-balls
		\begin{algorithmic}[1] 
			\STATE Treat the whole data set $D$ as a granular-ball $A_1^i$, where $i$ is initialized to 1, and represents the number of iterations;
			\STATE $balls$ were initialized to $A_1^1$;
			\STATE Randomly select $k$-1 points as initial division centers on $D$, and compute the distance from all points to the division centers;
			\REPEAT
			\FOR {each $A_j^i \in balls$}
			\STATE The purity $T_j^i$ is equal to the percentage of majority samples in $A_j^i$;
			\IF {$T_j^i < T$}
			\STATE Randomly select $k$-1 heterogeneous points as the new division centers, where $k$ represents the number of different labels in the balls;
			\STATE Compute the distances from the points in the ball to the new division centers;
			\STATE Based on the distances in step 9, granular-balls $A_1^{i+1},A_2^{i+1},\cdots,A_k^{i+1}$ are generated;
			\STATE $balls=balls-\{A_j^i\}$
			\STATE $balls=balls+\{A_1^{i+1}+A_2^{i+1}+\cdots+A_k^{i+1}\}$;
			\ENDIF	
			\ENDFOR
			\UNTIL {$\left| balls\right|$  does not increase}
			\STATE Perform a global division.
		\end{algorithmic}
	\end{algorithm}

	\subsection{Time Complexity}
	The time complexity of $k$-means is O($Nkt$)~\cite{38zhou2009novel}, where $k$ represents the number of clusters, and $t$ represents the number of iterations. The convergence speed of $k$-means is fast and can be considered approximately linear. The acceleration granular-ball generation method only needs to compute the distance between the data in this cluster and the new division centers each time when generating granular-balls. Assuming a data set with $m$ classes of data, in the first round of splitting, computation times are $mN$.
	
	In the second round, $m\left ( m-1 \right )$ new division centers are newly generated, and computation times are approximately
	
	\begin{equation}
		m\left ( m-1 \right )*\frac{N}{m}=\left ( m-1 \right )N.
	\end{equation}
	
	In the third round, $m^2\left ( m-1 \right )$ division centers are newly generated, and computation times is approximately
	
	\begin{equation}
		m^2\left ( m-1 \right )*\frac{N}{m^2}=\left ( m-1 \right )N.
	\end{equation}
	
	In the fourth round, $m^3\left ( m-1 \right )$ division centers are newly generated, and computation times is approximately
	
	\begin{equation}
		m^3\left ( m-1 \right )*\frac{N}{m^3}=\left ( m-1 \right )N.
	\end{equation}
	
	...	
	
	Assuming a total of $n$ iterations, the time complexity of the last global division is O($kN$), where $k$ is the number of granular-balls, and the total time complexity is O($(mn-n+k+1)N$). However, it is worth noting that the granular-balls will stop splitting when the splitting conditions are not met. Most granular-balls will stop splitting halfway through. Therefore, the actual time complexity of the acceleration granular-ball generation method is much lower than O($(mn-n+k+1)N$). The time complexity of the acceleration method is still linear, which avoids reduces unnecessary calculations.	
	\section{An Adaptive Granular-ball Generation Method}	
	\subsection{Motivation}
	Through the incoming purity threshold parameter, the existing method can generate granular-balls with meeting the purity threshold. The main problem of the existing method is that the purity threshold parameter cannot adapt to the data distribution of each data set, and it is difficult to find a splitting standard that matches the data distribution for each data sets. To solve this problem, we propose a purity adaptive granular-ball generation method based on the acceleration granular-ball generation method. The purity adaptation is important for the granular-ball generation, so that granular-ball generation process is completely parameter-free, and the completely parameter-free classifier, GB$k$NN, has been developed.
	
	\subsection{The Adaptive Conditions of Granular-ball Splitting}	
	
	In this section, we proposed three adaptive conditions to realize the adaptive generation of granular-balls including: whether the weighted purity sum of child balls for each granular-ball increases or not, whether there is overlap between any pair of heterogeneous granular-balls, and whether each granular-ball reaches to the lower bounder of purity, i.e., the purity of the initial granular-ball of the whole data set. The specific design are as follows in detail.
	
	\subsubsection{Weighed purity sum of child balls} \label{condition1}
	The purity is designed for measuring the quality of a granular-ball. So, a direct idea to design a indicator to measure the child balls' purity. Then, whether a granular-ball should be split or not is determined by whether its child granular-balls' purity becomes larger than itself. Considering the fact that the more samples in the ball, the more important the ball is, so we design the weighed purity sum of child balls for measuring the child balls' purity as shown in Definition \ref{def:wps}
	
	\begin{definition} \label{def:wps}
		Given granular-balls $A$ and its child balls $A_i(i=1,2,\dots,k)$, where $\bigcup_{i=1}^{k}{A_i}=A$ and $\bigcap_{i=1}^{k}{A_i}=\emptyset$. $\left| .\right| $ denotes the number of elements in a set. $k$ denotes the number of classes in $A$. $A_i^l $ denotes a set consisting of those samples whose label is equal to $l$, and $A_i^* $, i.e., $l=*$, represents the set consisting of those samples in the majority class in the $A_i$. The weighted purity sum $W$ of the child balls of $A$ can be defined as
		\begin{equation}
			\begin{aligned}
				W &=\frac{\left| A_1\right|}{\left| A\right|}\times\frac{\left| A_1^*\right|}{\left| A_1\right|}+\frac{\left| A_2\right|}{\left| A\right|}\times\frac{\left| A_2^*\right|}{\left| A_2\right|}+\cdots+\frac{\left| A_k\right|}{\left| A\right|}\times\frac{\left| A_k^*\right|}{\left| A_k\right|}\\
				&=\sum_{i=1}^{k}{\frac{\left| A_i^*\right|}{\left| A\right|}}\\
				&=\frac{\sum_{i=1}^{k}{\left| A_i^*\right|}}{\left| A\right|}.
			\end{aligned}
			\label{equ:w}
		\end{equation}
		\label{d1}
	\end{definition}
	
	Based on the Definition \ref{def:wps}, Theorem \ref{theom:wps} is proposed to describe the condition of that a granular-ball should be split.
	
	\newtheorem{theorem}{\textbf{Theorem}}
	\begin{theorem} \label{theom:wps}
		Given a granular-ball $A$, whose label is denoted by $label(A)$ and purity by $T=\frac{\left| A^*\right|}{\left| A\right|}$, and its child granular-ball $A_i(i=1,2,\dots,k)$, where $k$ is the number of the child granular-balls.  $label(A) = L$. $W$ represents the weighted purity sum of $A_i$.\par
		[1] $\forall A_i \subset A$, if $label\left ( A \right )=label\left ( A_i \right )$, then $W = T$;\par
		[2] $\exists A_i \subset A$, if $label\left ( A \right )\neq label\left ( A_i \right )$, then $W > T$.
	\end{theorem}
	\textbf{Proof:} 
	
	[1] When $\forall A_i \subset A$ and $label\left ( A \right )=label\left ( A_i \right )$, the majority samples in $A$ are also the majority samples in all child balls, so we have
	\begin{equation}
		A^{*} = A^{l},
		\label{equ:parent}
	\end{equation}
	\begin{equation}
		A_i^{*} = A_i^{l}.
		\label{equ:child}
	\end{equation}
	At the same time, the majority samples in $A$ are equal to the sum of the majority samples in all the child balls. Combining with the Equ. \ref{equ:parent} and Equ. \ref{equ:child}, we get
	\begin{equation}
		\sum_{i=1}^k{A_i^*}=\sum_{i=1}^k{A_i^l}=A^l=A^*.
		\label{equ:parent-child}
	\end{equation}
	From the Equ. \ref{equ:parent-child} and the Definition \ref{def:wps}, we can easily get
	\begin{equation}
		W=\frac{\sum_{i=1}^k{\left| A_i^*\right| }}{\left| A\right|}=\frac{\left| A^*\right|}{\left| A\right|} =T.
		\label{equ:pd}
	\end{equation}
	So,
	\begin{equation}
		W = T.
	\end{equation}
	
	[2] When $\exists A_i\subset A$ and $label\left ( A \right )\neq label\left ( A_i \right )$, the majority sample in $A$ are not necessarily the majority sample in all child balls. Assuming $A_m(m=1,2,\dots,k)$ has a different label with $A$, we can get
	\begin{equation}
		\left| A_m^*\right| >\left|A_m^l \right|. 
		\label{equ:Am*>Aml}
	\end{equation}
	In addition, the samples labeled $l$ in the parent ball equal to the sum of those in all child balls, we have
	\begin{equation}
		\left| A^l\right| =\sum_{i=1}^{k}\left|A_i^l \right|. 
		\label{equ:al}
	\end{equation}
	Combining with Equ. \ref{equ:Am*>Aml} and Equ. \ref{equ:al}, we get
	\begin{equation}
		\begin{aligned}
			\sum_{i=1}^k{\left| A_i^*\right| } &=\left| A_1^*\right|+\left| A_2^*\right|+\dots+\left| A_m^*\right|+\dots+\left| A_k^*\right| \\
			&>\left| A_1^l\right|+\left| A_2^l\right|+\dots+\left| A_m^l\right|+\dots+\left| A_k^l\right| \\
			&=\sum_{i=1}^k{\left| A_i^l\right| }  \\
			&=\left| A^l\right|\\
			&=\left| A^*\right|.
		\end{aligned}
		\label{equ:Ai*>Ail}
	\end{equation}
	From the Equ. \ref{equ:Ai*>Ail} and the Definition \ref{def:wps}, we can easily get
	\begin{equation}
		W=\frac{\sum_{i=1}^k{\left| A_i^*\right| }}{\left| A\right|}>\frac{\left| A^*\right|}{\left| A\right|} =T.
		\label{equ:pd}
	\end{equation}
	So,
	\begin{equation}
		W > T.
	\end{equation}
	
	When $W$ is greater than $T$, it means the label of some child balls is different from the parent ball, i.e., the minority samples in the parent ball become the majority samples in some child balls. It can be concluded that the weighted purity sum of the child balls will be greater than the purity of the parent ball, and the number of correctly classified samples will increase.
	When $W$ equals $T$, it represents a special case that the parent ball and all child balls have the same label.  
	
	\begin{figure}[!t]
		\centering
		\subfigure[]{\includegraphics[width=0.23\textwidth]{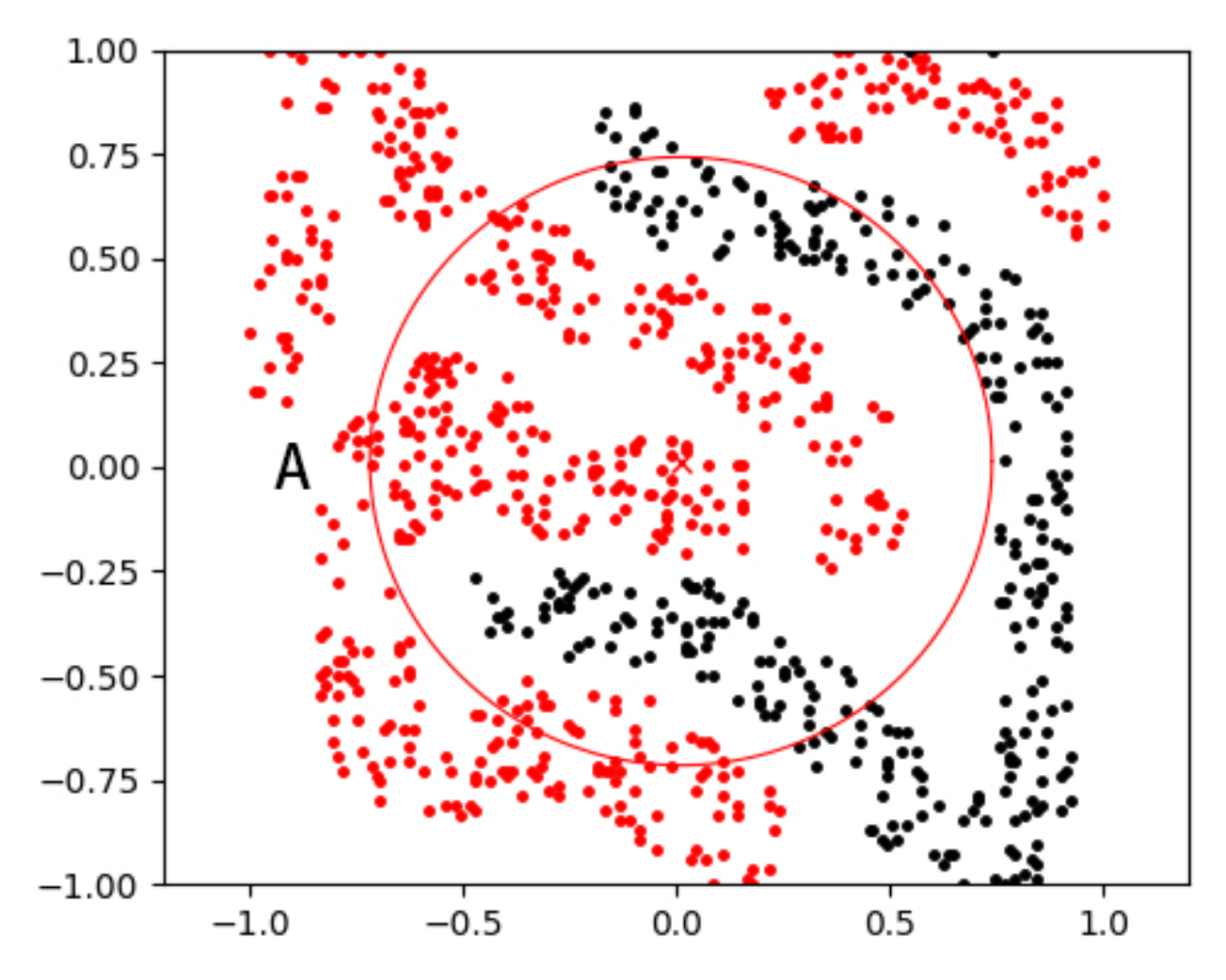}}
		\subfigure[]{\includegraphics[width=0.23\textwidth]{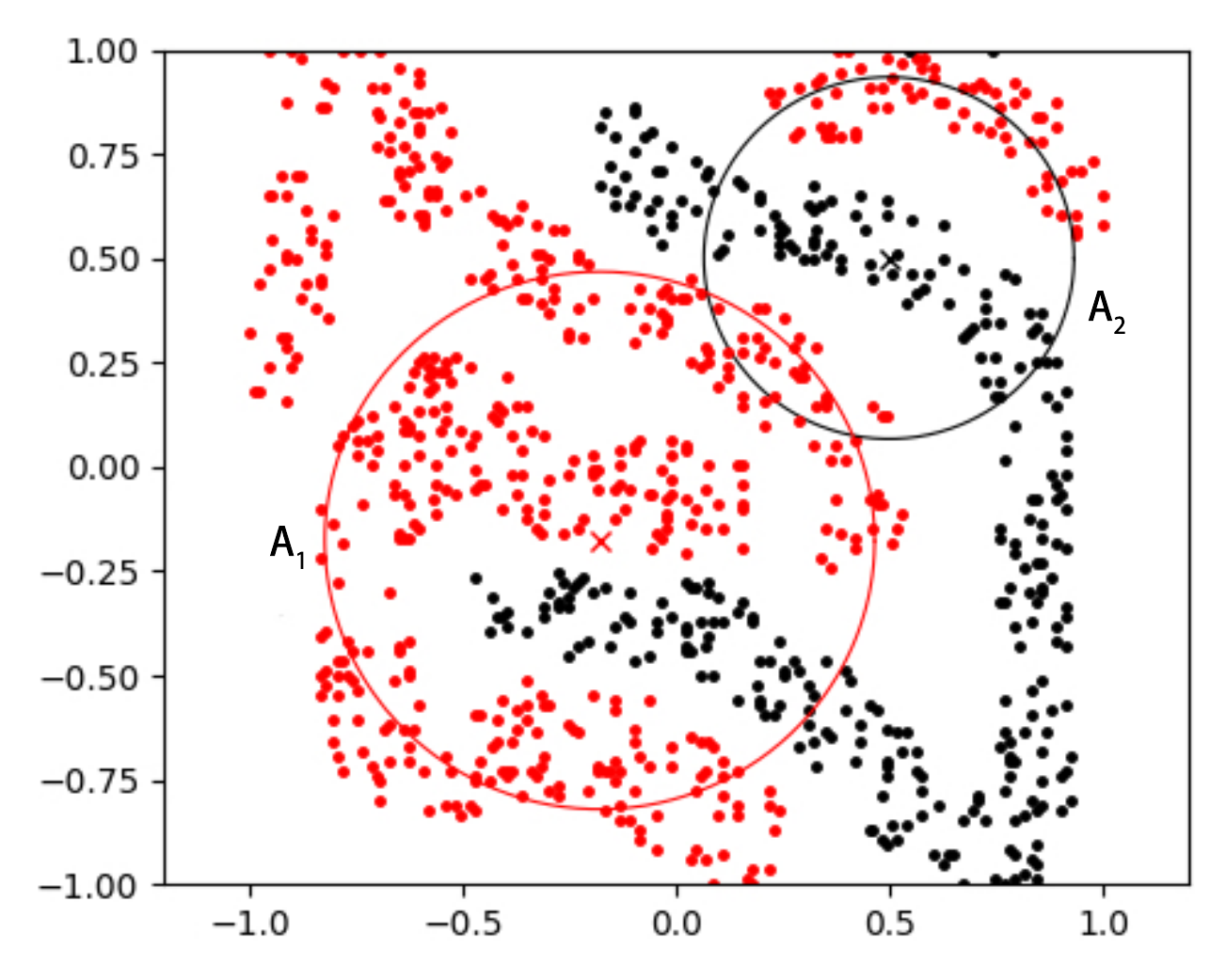}}
		\subfigure[]{\includegraphics[width=0.23\textwidth]{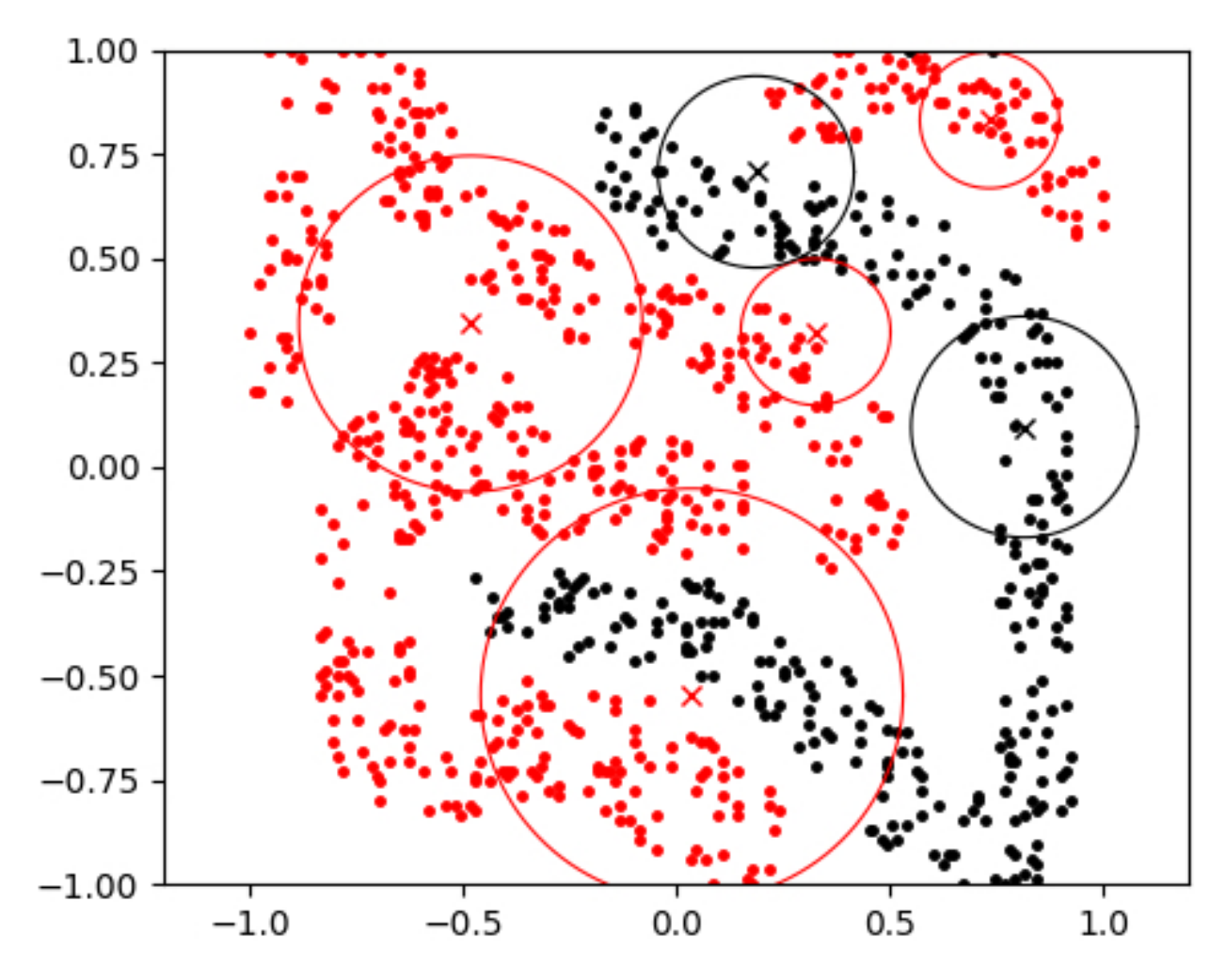}}
		\subfigure[]{\includegraphics[width=0.23\textwidth]{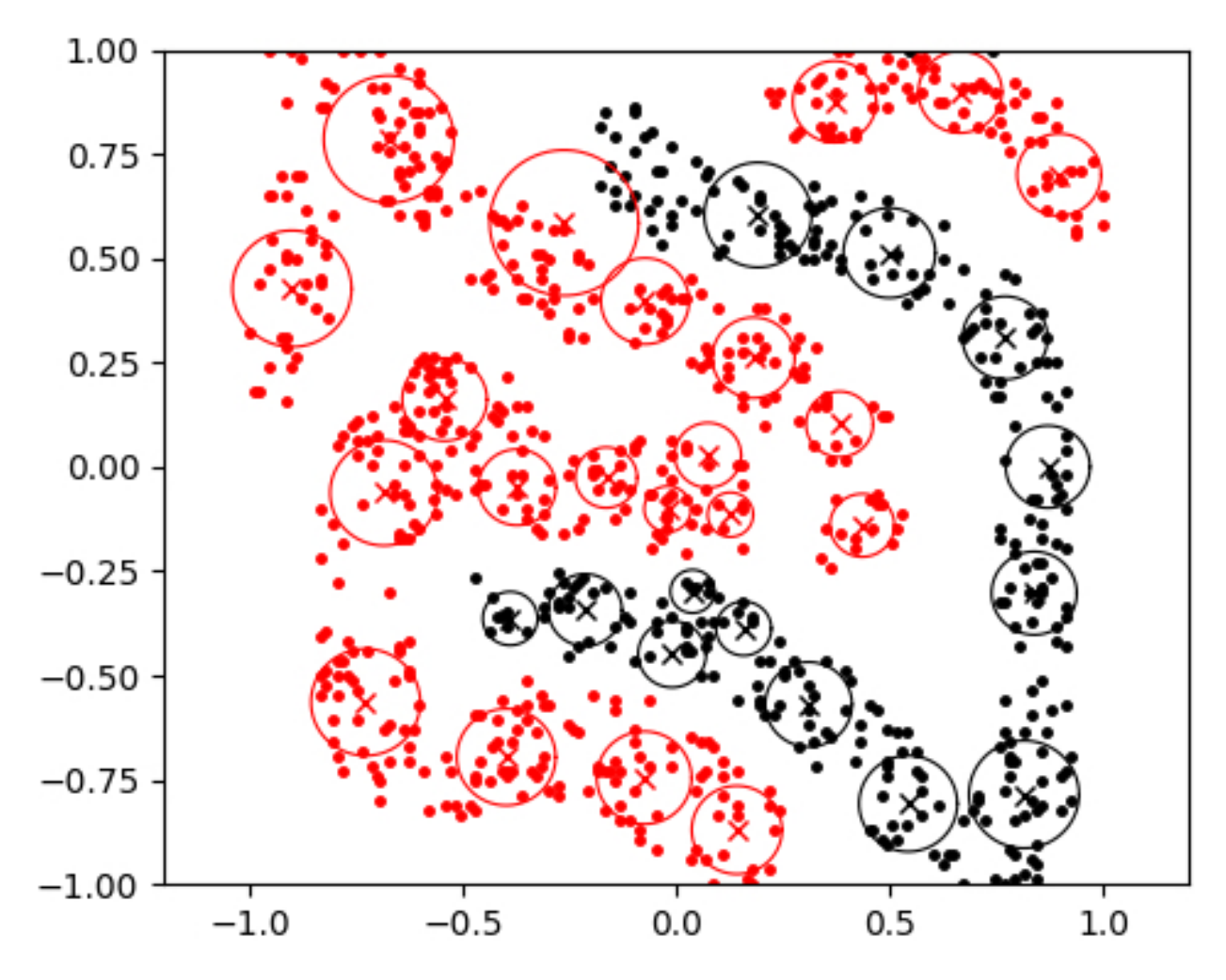}}
		\subfigure[]{\includegraphics[width=0.23\textwidth]{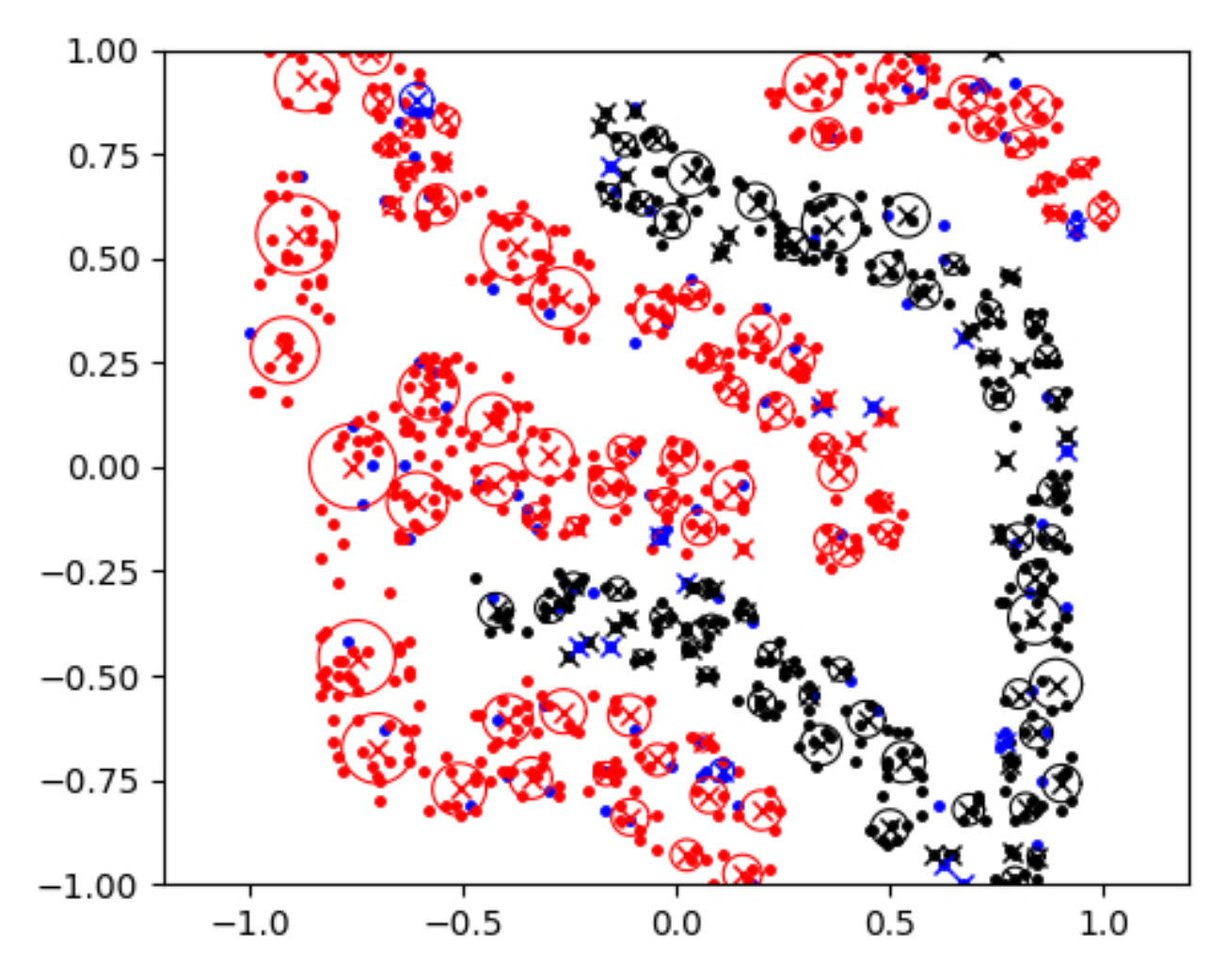}}
		\centering{\caption{The situation before and after splitting using the adaptive granular-ball generation method. (a) The parent ball before splitting; (b) The child balls after splitting; (c) The situation after de-overlap; (d) The algorithm convergence result using the proposed adaptive method; (e) The experimental result using the proposed adaptive method in a noisy data set where the noise points are colored with blue.}
			\label{Fig9}}
		\vspace{-1em}
	\end{figure}

	As shown in Fig. \ref{Fig9}, the ball $A$ in (a) is split into ball $A_1$ and ball $A_2$ in (b) using Theorem \ref{theom:wps}, and the labels of $A$ and $A_2$ are different. At this time, the minority class in $A$ becomes the majority of classes in $A_2$ , so the weighted purity sum of $A_1$ and $A_2$ will be greater than the purity of $A$. From the perspective of GB$k$NN, the number of samples with correct classification also increases.	But in this case, there will still be premature convergence, because the accuracy does not increase monotonically. Fig. \ref{Fig9}(b) shows a simple example of premature convergence only using the conditions in Section \ref{condition1}, that is, there is overlap between heterogeneous granular-balls.
	
	To this end, we introduce the second condition that there can be no overlap between heterogeneous granular-balls. 
	
	\subsubsection{De-overlap between heterogeneous granular-balls}  \label{condition2}
	For the problem of granular-balls overlap, it is necessary to further detect whether there is heterogeneous granular-balls on the basis of Section \ref{condition1}, and further split and refine the overlapping granular-balls to make the decision boundary clearer. In order to improve efficiency, the next round of overlap detection only needs to traverse the child granular-balls of the granular-balls that have overlapped. The boundary overlap problem of heterogeneous granular-balls can be defined as the following model:
	
	\begin{equation}
		\begin{aligned}
			&\exists \left \| c_i-c_j \right \|\leq \left \| r_i+r_j \right \|,for\ i,j\in\left \{ 1,2,..., m \right \}\\
			&s.t.\ label\left ( GB_i \right )\neq label\left ( GB_j \right ).\\
		\end{aligned}
	\end{equation}
	
	Among them, $c_i$ represents the center of the $i^{th}$ granular-ball, $r_i$ represents the radius of the $i^{th}$ granular-ball, and $m$ is the total number of granular-balls. In addition, the effect of the condition ``$label\left ( GB_i \right )\neq label\left ( GB_j \right )$" is to concentrate the boundary overlap problem between heterogeneous granular-balls, and reducing the cost of computing and analysis of the problem. The overlap between the same kind of granular-balls will not affect the decision boundary. As shown in Fig. \ref{Fig9}(c), it can be seen that, compared to Fig. \ref{Fig9}(b), the granular-balls after de-overlap are more suitable for the data distribution.
	
	\subsubsection{An adaptive purity lower bound}  \label{condition3}
	
	In addition, the purity of the granular-balls should has an adaptive lower bound. The lower bound is the proportion of the initial majority of samples of the total sample, that is, the purity of the initial granular-ball. As shown in Fig. \ref{Fig9}(c), the granular-balls will adaptively generate granular-balls with a lower purity than the initial purity. The quality of such granular-balls is too low, which reduces the classification accuracy of the final GB$k$NN. For minority samples, the proportion of incorrectly classified samples can be considered as the noise rate. That is, the purity of all granular-balls must be greater than the purity of the initial granular-ball.
	
	Fig. \ref{Fig9}(d) shows the result of granular-ball generation using the adaptive method. The two colored granular-balls in the figure represent two classes of data. In addition, Fig. \ref{Fig9}(e) shows the result of granular-ball generation using the adaptive method when some label noise points are added. The blue points in the figure is represented by noise data, and the other two colored points represent the two original data. The noise points are generated by randomly changing the labels of samples in the data set. The optimization goal of the adaptive granular-ball generation method can be expressed as

	\begin{scriptsize}
		\begin{equation}
			\begin{aligned}
				&Min\ \lambda _1*n/\sum_{j=1}^{m} (GB_j)+\lambda _2*m,\\
				&s.t.\ quality(GB_j) > T_0, W(GB_j^{'}) > quality(GB_j),\\
				&\left \| c_i-c_j \right \|> \left \| r_i+r_j \right \|( i,j\in \left [ 1,m \right ],\ label(GB_i)\neq label(GB_j)),\\
			\end{aligned}
		\end{equation}	
	\end{scriptsize}
	
	where $\lambda _1$ and $\lambda _2$ are the corresponding weight coefficients and $k < n$, and $c_i, r_i$ represents the center and radius of $GB_i$ respectively. $T_0$ denote the adaptive purity lower bound of granular-balls, and $GB_j^{'}$ represents the child granular-balls of $GB_j$. In addition, $label(GB_i)$ and $W$ are mentioned above.
	
	\subsection{Method Design}
\begin{figure}[htbp!]
		\centering
		\fbox{\includegraphics[width=0.45\textwidth]{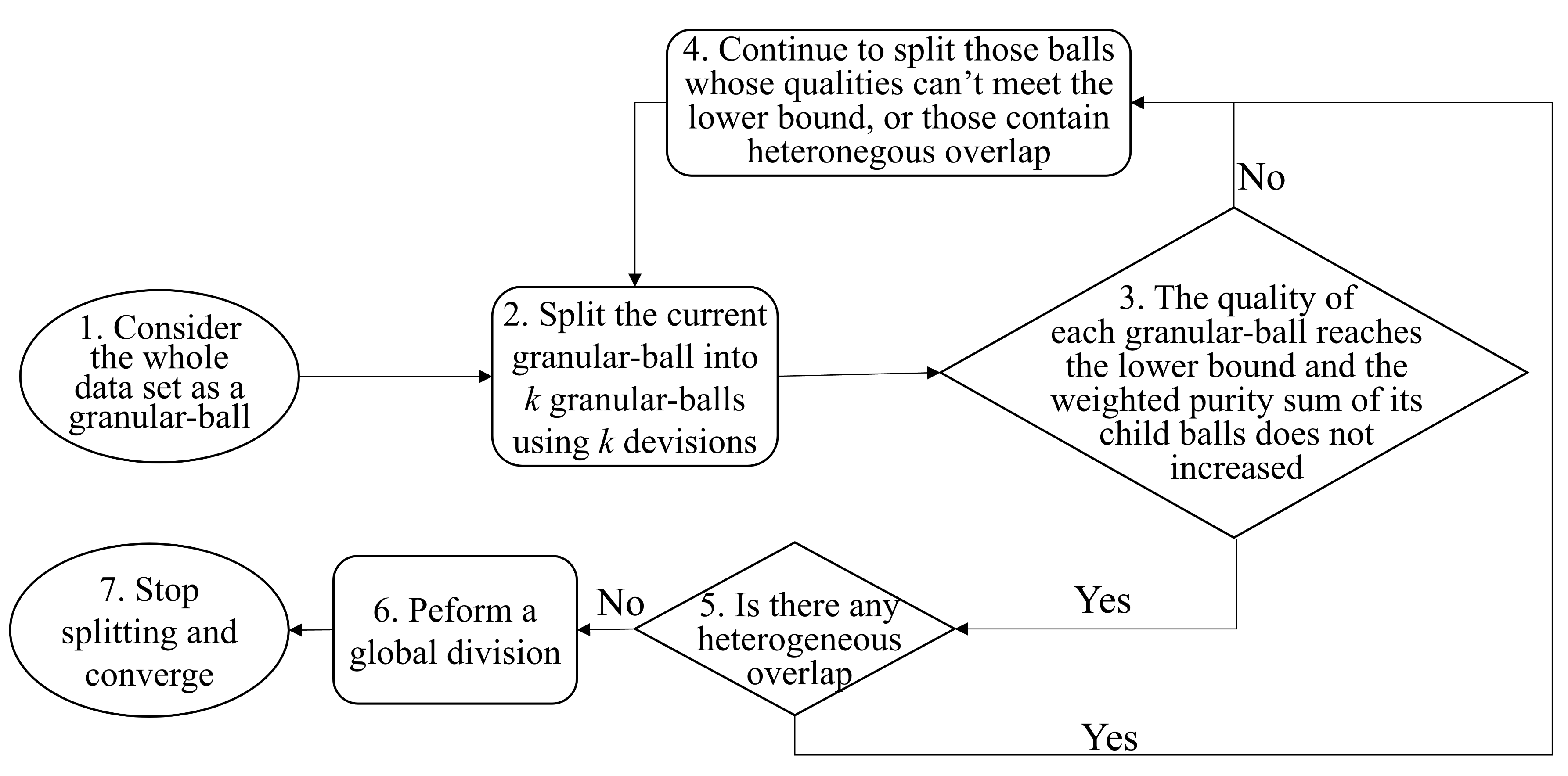}}
		\centering{\caption{The basic idea of the adaptive granular-ball generation method.}
		\label{Fig10}}
		\vspace{-1em}
	\end{figure}

	The basic idea of granular-ball generation of the adaptive granular-ball generation method is shown in Fig. \ref{Fig10}. 
	
	In step 2 in Figure \ref{Fig10}, based on the accelerated granular-ball generation method, k division is used to split the granular-ball, where $k$ denotes the number of classes in a granular-ball. So, the computation cost is decreased directly. In addition, as shown in step 3, when the weighted purity sum of the child balls is greater than the purity of its parent ball and the purity of the granular-ball reaches the lower bound, the child balls are retained and whether there is overlap between heterogeneous granular-balls is detected. As shown in Fig. \ref{Fig9}(d), the boundary of the granular-balls when the algorithm converges is very consistent with that of the data set. 
	
	The algorithm design for adaptive granular-ball generation method is as Algorithm \ref{alg2}.
	
	\begin{algorithm}[htbp!]
		\caption{Granular-ball generation adaptive method}
		\label{alg2}
		\textbf{Input}: Data set $D$\\
		\textbf{Output}: The granular-balls
		\begin{algorithmic}[1] 
			\STATE Treat the whole data set $D$ as a granular-ball $A_1^i$, where $i$ is initialized to 1, and represents the number of iterations;
			\STATE $balls$ were initialized to $A_1^1$;
			\REPEAT
			\FOR {each $A_j^i \in balls$}
			\STATE Implement the acceleration granular-ball generation method on $A_j^i$, pre-generate $k$ granular-balls $A_1^{i+1},A_2^{i+1},\cdots, A_k^{i+1}$;
			\STATE The $T_{j}^{i}$ represents the purity of $A_j^i$;
			\STATE The $W_{j}^{i}$ represents the weighted purity sum of the child balls of $A_j^i$;
			\IF {$W_{j}^{i} > T_{j}^{i}$ or $T_{j}^{i} <= T_1^1$}
			\STATE $balls=balls-\{A_j^i\}$
			\STATE $balls=balls+\{A_1^{i+1}+A_2^{i+1}+\cdots+A_k^{i+1}\}$;			
			\ENDIF
			\ENDFOR
			\STATE De-overlap between heterogeneous granular-balls;
			\UNTIL {$\left| balls\right|$  does not increase}
			\STATE Perform a global division.
		\end{algorithmic}
	\end{algorithm}

	\section{Experiments}
	
	To demonstrate the feasibility and effectiveness of the acceleration granular-ball generation method and the adaptive granular-ball generation method, we compared them with $k$NN and two popular or the state-of-the-art methods based on granular computing, including GB$k$NN~\cite{17xia2019granular} and GBS~\cite{37xia2021granular}. Because of the robustness of the granular-ball, our experiments are carried out both on the raw data sets and the noise data sets. We verifies the performance on accuracy of the acceleration granular-ball generation method and the adaptive granular-ball generation method, and on efficiency of the acceleration method. We randomly selected ten real data sets from UCI benchmark data sets as shown in the following tables. Experimental hardware environment: PC with an Intel Core i7-107000 CPU @2.90 GHz with 32 G RAM. Experimental software environment: Python 3.9. 
	
	\subsection{Experiments on Raw Data Sets}
	In this section, we split the data set into ten parts, take one part for testing, and use the test accuracy as the evaluation index to verify the effectiveness of the acceleration granular-ball generation method and the adaptive granular-ball generation method. Since granular-ball generation still has a certain randomness, we do experiments on each method ten times, and take the average classification accuracy of the ten experiments results for comparison. The $k$NN method use the ten-fold cross-validation result.	
	
	Table \ref{tab1} shows the experimental accuracy of $k$NN under noise-free conditions. ``Acc$^+$", ``Adp", ``Origin", and $k$NN represent the acceleration granular-ball generation method, the adaptive granular-ball generation method, the existing granular-ball generation method, and $k$NN, respectively. ``mean" and ``max" represent the experimental accuracy of the average distance and the maximum distance as the radius of the granular-ball. The proposed two methods, the acceleration granular-ball generation method and the adaptive granular-ball generation method, obtained a better performance on accuracy compared to the existing method and $k$NN. The decision boundary obtained by using the existing methods is still not clear enough. Therefore, when measuring the $k$NN accuracy, the granular-ball closest to the test point is likely to be inaccurate. The global division is performed after the splitting of the granular-ball is stopped. Therefore, the two methods proposed by us can obtain higher $k$NN accuracy than the existing methods in the raw data set.

	\begin{table}[!htbp]
		\centering
		\caption{Comparison of average test accuracy (raw data sets)}
		\label{tab1}
		\setlength{\tabcolsep}{1.6mm}{
			\begin{tabular}{cccccccc}
				\toprule
				\multirow{2}{*}{Data} & \multicolumn{2}{c}{Acc$^+$} & \multicolumn{2}{c}{Adp} & \multicolumn{2}{c}{Origin} & \multirow{2}{*}{$k$NN} \\
				& mean           & max           & mean      & max      & mean         & max         &                      \\ \toprule
				fourclass                  & 0.990                      & 0.987                     & 0.988                         & 0.973                        & 0.990                     & 0.958                    & \textbf{0.999}           \\
				svmguide1                  & 0.959                      & 0.965                     & 0.930                         & \textbf{0.970}               & 0.960                     & 0.796                    & 0.960                    \\
				diabetes                   & 0.824                      & \textbf{0.838}            & 0.834                         & 0.824                        & 0.718                     & 0.697                    & 0.748                    \\
				breastcancer               & 0.950                      & 0.972                     & 0.977                         & 0.965                        & \textbf{0.982}            & 0.962                    & 0.973                    \\
				creditApproval             & \textbf{0.770}             & 0.769                     & 0.722                         & 0.714                        & 0.669                     & 0.599                    & 0.659                    \\
				votes                      & 0.906                      & 0.917                     & 0.868                         & \textbf{0.922}               & 0.871                     & 0.722                    & 0.875                    \\
				svmguide3                  & \textbf{0.831}             & 0.828                     & 0.812                         & 0.818                        & 0.786                     & 0.776                    & 0.788                    \\
				sonar                      & \textbf{0.895}             & 0.886                     & 0.833                         & 0.855                        & 0.833                     & 0.757                    & 0.831                    \\
				splice                     & 0.745                      & \textbf{0.807}            & 0.765                         & 0.796                        & 0.605                     & 0.595                    & 0.681                    \\
				mushrooms                  & 0.994                      & \textbf{1.000}            & \textbf{1.000}                & \textbf{1.000}               & 0.993                     & 0.715                    & \textbf{1.000}           \\
				Average                    & 0.886                      & \textbf{0.897}            & 0.873                         & 0.884                        & 0.841                     & 0.758                    & 0.851                   \\ \toprule
			\end{tabular}
		}
	\end{table}
	
	\begin{table}[!htbp]
		\centering
		\caption{Comparison of average test accuracy after sampling with GBS (raw data sets)}
		\label{tab2}
		\setlength{\tabcolsep}{2.2mm}{
			\begin{tabular}{cccccc}
				\toprule
				Data           & Origin-GBS      & Acc$^+$-GBS     & Adp-GBS         & $k$NN      \\ \toprule
				fourclass      & 0.9890          & 0.9902          & 0.9942          & \textbf{0.9971} \\
				svmguide1      & 0.9558          & \textbf{0.9612} & 0.9587          & 0.9596          \\
				diabetes       & 0.7331          & \textbf{0.7494} & 0.7448          & 0.7312          \\
				breastcancer   & 0.9644          & 0.9585          & \textbf{0.9696} & 0.9644          \\
				creditApproval & \textbf{0.6855} & 0.6725          & 0.6609          & 0.6623          \\
				votes          & 0.8884          & 0.9000          & \textbf{0.9029} & 0.8870          \\
				svmguide3      & 0.7835          & 0.7803          & \textbf{0.7863} & 0.7807          \\
				sonar          & 0.8048          & \textbf{0.8476} & 0.8262          & 0.8048          \\
				splice         & 0.6964          & \textbf{0.7265} & 0.7061          & 0.6750          \\
				mushrooms      & 0.9994          & \textbf{1.0000} & \textbf{1.0000} & \textbf{1.0000} \\
				Average        & 0.8500          & \textbf{0.8586} & 0.8550          & 0.8462          \\ \toprule
			\end{tabular}
		}
	\end{table}
	
	The column 1-3 in Table \ref{tab2} is based on the GBS method. Firstly, we use the first three methods in Table \ref{tab2} to generate the granular-balls; and then the GBS method is used to sample the generated granular-balls; finally, $k$NN is used to classify the sampled result. The last column represents directly classifying the raw data set with $k$NN. According to the paper\cite{37xia2021granular}, the purity is set from 0.54 to 1.0 with the step size of 0.2 GBS algorithm. It can be seen that our two methods have a higher accuracy on most data sets than the other two.
	
	\begin{table*}[!htbp]
		\centering
		\caption{Comparison of the running time of the acceleration granular-ball generation method and the existing method}
		\label{tab3}
		\includegraphics[width=1\textwidth]{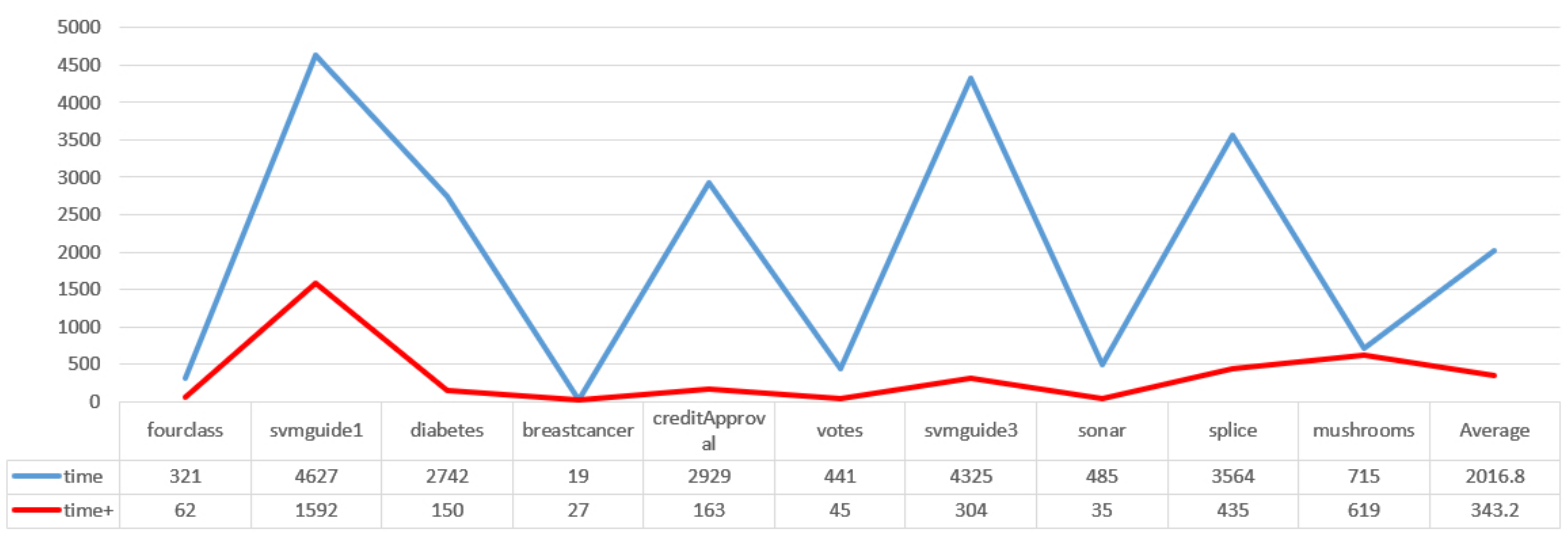}
		\hfil
		\centering
		\caption{Comparison of the number of granular-balls generated by the acceleration granular-ball generation method and the existing method}
		\label{tab4}
		\setlength{\tabcolsep}{2.1mm}{
			\begin{tabular}{cccccccccccc}
				\toprule
				Data                    & fourclass   & svmguide1    & diabetes     & breastcancer & creditApproval & votes       & svmguide3    & sonar       & splice       & mushrooms   & Average      \\ \toprule
				balls                    & \textbf{31} & 533          & 394          & \textbf{2}   & 426            & 61          & 597          & \textbf{69} & 517          & \textbf{14} & 264          \\
				balls+                   & \textbf{31} & \textbf{390} & \textbf{348} & 15           & \textbf{364}   & \textbf{60} & \textbf{524} & 73          & \textbf{506} & 39          & \textbf{235} \\ \toprule
			\end{tabular}
			\hfil
		}
	\end{table*}

	In order to show the efficiency of the acceleration granular-ball generation method, we choose the existing granular-ball generation method as the comparison method.	Table \ref{tab3} shows the running time of the two methods on raw data sets. The ``time+" and ``time" denote the acceleration method and the the existing granular-ball generation method respectively. Table \ref{tab4} shows the comparison of the number of granular-balls generated by the acceleration method and the existing method on raw data sets, where ``ball+", ``ball" denote the acceleration method and the existing method respectively. Compared with existing granular-ball generation method from Table \ref{tab1}-\ref{tab4}, the acceleration method has a higher accuracy and efficiency on most data sets, while generating the similar number of granular-balls.

	\subsection{Experiments on Noise Data Sets}
	In this section, each data set has four class noise rates, namely 10\%, 20\%, 30\% and 40\%. Noise is generated by changing the labels of randomly selected samples in a data set. Tables \ref{tab5}-\ref{tab8} show the GBS highest average test accuracy obtained from the purity optimization of the existing method under different noise rates, and the GBS highest average accuracy of adaptive granular-ball generation method and acceleration granular-ball generation method. The purity is also set from 0.54 to 1.0 with the step size of 0.2 in GBS algorithm. The acceleration method and the adaptive method adopt the strategy of selecting heterogeneous sample points as the new clustering centers when splitting a granular-ball. This can make the algorithm converge faster, but it will reduce the accuracy when dealing with noisy data sets. It can also be seen from the experimental results of noise data sets that the acceleration granular-ball generation method and the adaptive granular-ball generation method can obtain a similar law to the existing granular-ball generation method on the noisy data, that is, when the noise rate in the data set is larger, the advantage to the original $k$NN is more obvious. However, it can also be seen that the adaptive method still show slightly lower accuracy than the existing method when dealing with the noise data sets. It is possible that the adaptive purity lower bound of granular-ball in the adaptive method is too low, so that some granular-balls with poor quality are generated, which affects the overall accuracy. However, the adaptive method significantly improves the existing method to make it adaptive.
	
	\begin{table}[!htbp]
		\centering
		\caption{Comparison of average test accuracy after sampling with GBS (noise rate 10\%)}
		\label{tab5}
		\setlength{\tabcolsep}{2.1mm}{
			\begin{tabular}{cccccc}
				\toprule
				Data           & Origin-GBS      & Acc$^+$-GBS     & Adp-GBS         & $k$NN      \\ \toprule
				fourclass      & \textbf{0.8815} & 0.8792          & 0.8763 & 0.8769          \\
				svmguide1      & 0.8523          & \textbf{0.8625} & 0.8461 & 0.8428          \\
				diabetes       & 0.6721          & \textbf{0.6935} & 0.6701 & 0.6578          \\
				breastcancer   & \textbf{0.8711} & 0.8504          & 0.8593 & 0.8393          \\
				creditApproval & \textbf{0.6442} & 0.6225          & 0.6283 & 0.6123          \\
				votes          & \textbf{0.8188} & 0.8029          & 0.7957 & 0.7957          \\
				svmguide3      & \textbf{0.7297} & 0.7285          & 0.7088 & 0.6964          \\
				sonar          & \textbf{0.7571} & 0.7357          & 0.7452 & \textbf{0.7571} \\
				splice         & 0.6449          & \textbf{0.6485} & 0.6388 & 0.6173          \\
				mushrooms      & \textbf{0.8926} & 0.8781          & 0.8358 & 0.8740          \\
				Average        & \textbf{0.7764} & 0.7702          & 0.7604 & 0.7570          \\ \toprule
			\end{tabular}
		}
	\end{table}
	
	\begin{table}[!htbp]
		\centering
		\caption{Comparison of average test accuracy after sampling with GBS (noise rate 20\%)}
		\label{tab6}
		\setlength{\tabcolsep}{2.1mm}{
			\begin{tabular}{cccccc}
				\toprule
				Data           & Origin-GBS      & Acc$^+$-GBS     & Adp-GBS         & $k$NN      \\ \toprule
				fourclass      & 0.7370          & \textbf{0.7948} & 0.7583          & 0.7046 \\
				svmguide1      & 0.7602          & \textbf{0.7677} & 0.7098          & 0.7156 \\
				diabetes       & 0.6214          & \textbf{0.6390} & 0.6065          & 0.5994 \\
				breastcancer   & 0.7652          & \textbf{0.7941} & 0.7815          & 0.6852 \\
				creditApproval & \textbf{0.6210} & 0.5775          & 0.5695          & 0.5696 \\
				votes          & \textbf{0.7217} & 0.7072          & 0.7014          & 0.6725 \\
				svmguide3      & 0.6663          & \textbf{0.6775} & 0.6261          & 0.6108 \\
				sonar          & \textbf{0.6786} & 0.6238          & 0.6452          & 0.6786 \\
				splice         & \textbf{0.5929} & 0.5857          & 0.5740          & 0.5699 \\
				mushrooms      & \textbf{0.7830} & 0.7584          & 0.6985          & 0.7314 \\
				Average        & \textbf{0.6947} & 0.6926          & 0.6671          & 0.6537 \\ \toprule
			\end{tabular}
		}
	\end{table}
	
	\begin{table}[!htbp]
		\centering
		\caption{Comparison of average test accuracy after sampling with GBS (noise rate 30\%)}
		\label{tab7}
		\setlength{\tabcolsep}{2.1mm}{
			\begin{tabular}{cccccc}
				\toprule
				Data           & Origin-GBS      & Acc$^+$-GBS     & Adp-GBS & $k$NN      \\ \toprule
				fourclass      & \textbf{0.6711} & 0.6659          & 0.6653  & 0.6156 \\
				svmguide1      & 0.6543          & \textbf{0.6809} & 0.6025  & 0.6019 \\
				diabetes       & 0.5669          & \textbf{0.5877} & 0.5370  & 0.5266 \\
				breastcancer   & \textbf{0.7052} & 0.6800          & 0.6467  & 0.6178 \\
				creditApproval & 0.5667          & \textbf{0.5688} & 0.5355  & 0.5355 \\
				votes          & \textbf{0.6435} & 0.6362          & 0.6232  & 0.5928 \\
				svmguide3      & \textbf{0.6116} & 0.6096          & 0.5735  & 0.5434 \\
				sonar          & 0.5690          & \textbf{0.5786} & 0.5667  & 0.5690 \\
				splice         & \textbf{0.5592} & 0.5500          & 0.5342  & 0.5245 \\
				mushrooms      & \textbf{0.6881} & 0.6495          & 0.5921  & 0.6151 \\
				Average        & \textbf{0.6236} & 0.6207          & 0.5877  & 0.5742 \\ \toprule
			\end{tabular}
		}
	\end{table}
	
	\begin{table}[!htbp]
		\centering
		\caption{Comparison of average test accuracy after sampling with GBS (noise rate 40\%)}
		\label{tab8}
		\setlength{\tabcolsep}{2.1mm}{
			\begin{tabular}{cccccc}
				\toprule
				Data           & Origin-GBS      & Acc$^+$-GBS     & Adp-GBS         & $k$NN      \\ \toprule
				fourclass      & \textbf{0.5775} & 0.5468          & 0.5740          & 0.5312 \\
				svmguide1      & 0.5711          & \textbf{0.5807} & 0.5340          & 0.5316 \\
				diabetes       & 0.5117          & \textbf{0.5468} & 0.5026          & 0.4890 \\
				breastcancer   & \textbf{0.5904} & 0.5778          & 0.5615          & 0.5230 \\
				creditApproval & 0.5080          & \textbf{0.5435} & 0.5145          & 0.4819 \\
				votes          & \textbf{0.5841} & 0.5652          & 0.5362          & 0.5319 \\
				svmguide3      & \textbf{0.5627} & 0.5498          & 0.5171          & 0.5104 \\
				sonar          & 0.4881          & 0.5405          & \textbf{0.5690} & 0.4833 \\
				splice         & \textbf{0.5321} & 0.5179          & 0.5184          & 0.5281 \\
				mushrooms      & \textbf{0.5860} & 0.5641          & 0.5255          & 0.5338 \\
				Average        & \textbf{0.5511} & 0.5533          & 0.5353          & 0.5144 \\ \toprule
			\end{tabular}
		}
	\end{table}
	
	\section{Conclusions and Future Work}
	This paper proposes a method for accelerating the granular-ball generation, which can greatly improve the efficiency of granular-ball generation while ensuring accuracy. At the same time, a new granular-ball clustering method is proposed, that is, the adaptive granular-ball generation method. This adaptive method avoids the problem that the existing method needs to manually set the purity threshold parameter, and makes the generation process of the granular-balls completely adaptive. Experiments show that the acceleration method has better performance than the adaptive method for both noisy and non-noised data. At the same time, as shown by experiments, experimental accuracy of the adaptive method is slightly lower than the existing method. It proves that our method is effective, but whether there are other adaptive methods, such as based on the consistency of the internal distribution of granular-balls, may develop a more effective granular-ball adaptive optimization method. However, the proposed methods exhibit lower accuracy in some cases than the existing method, so we will study how to improve their accuracy in the future work. 
	
	\section{Acknowledgments}
	This work was supported in part by the National Natural Science Foundation of China under Grant Nos. 62176033 and 61936001, National Key Research and Development Program of China under Grant No. 2019QY(Y)0301, the Natural Science Foundation of Chongqing under Grant No. cstc2019jcyj-cxttX0002 and by NICE: NRT for Integrated Computational Entomology, US NSF award 1631776.
	
	\ifCLASSOPTIONcaptionsoff
	\newpage
	\fi
	
	\bibliographystyle{unsrt}
	\bibliography{AGB}
	
	\ifCLASSOPTIONcaptionsoff
	\newpage
	\fi
	
	\begin{IEEEbiography}[{\includegraphics[width=1in,height=1.25in,clip,keepaspectratio]{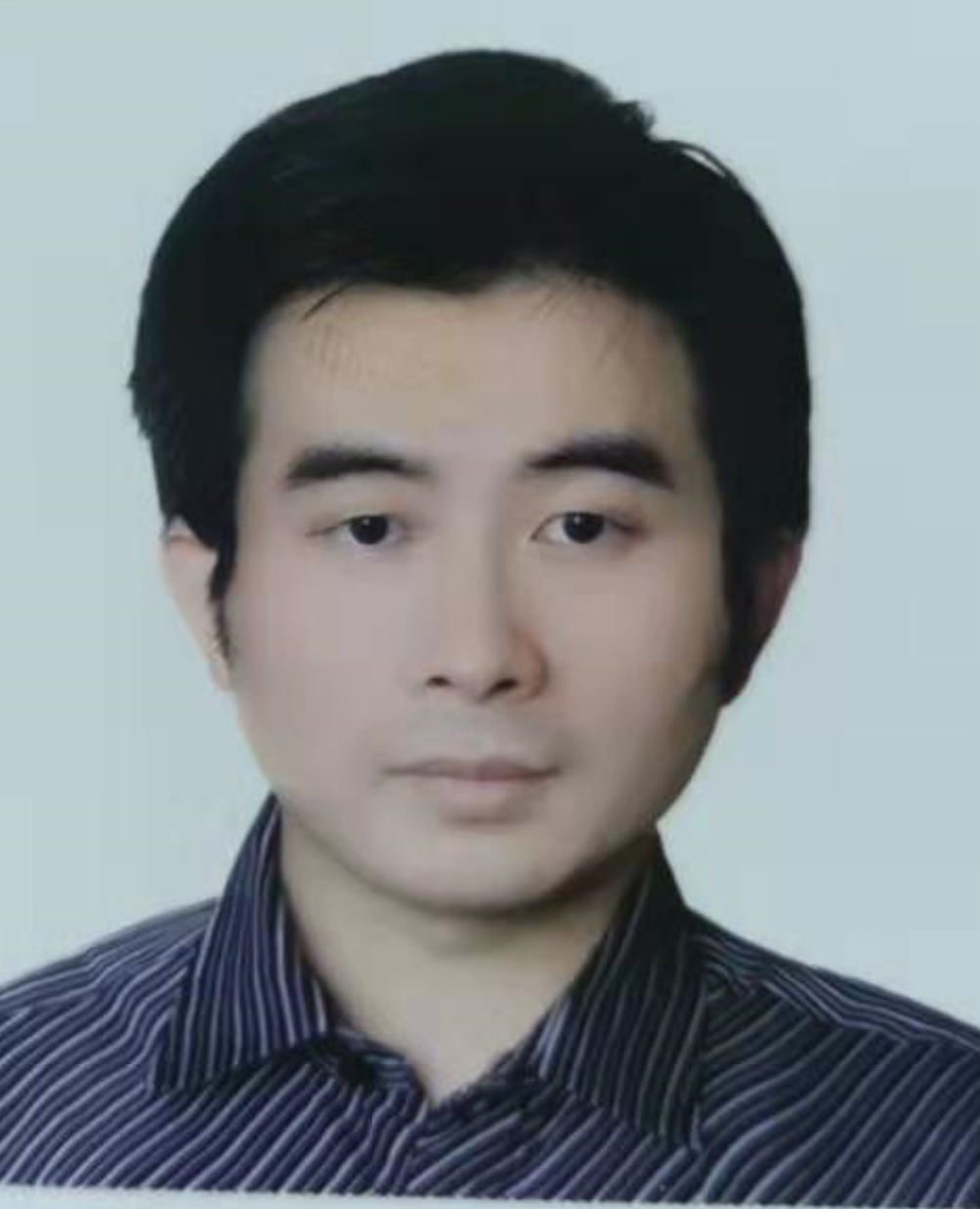}}]{Shuyin Xia} received his B.S. degree and M.S. degree in computer science in 2008 and 2012, respectively, from Chongqing University of Technology in China. He received his Ph.D. degree from the College of Computer Science, Chongqing University in China. He is an IEEE Member. He is currently an associate professor at the College of Computer Science and Technology, Chongqing University of Posts and Telecommunications. He is also the executive deputy director of the Big Data and Network Security Joint Lab of CQUPT. He has published more than 30+ papers in journals and conferences, including IEEE T-PAMI, T-KDE, T-NNLS, T-CYB and Information Science. His research interests include classifiers and granular computing.
	\end{IEEEbiography}
	
	\begin{IEEEbiography}[{\includegraphics[width=1in,height=1.25in,clip,keepaspectratio]{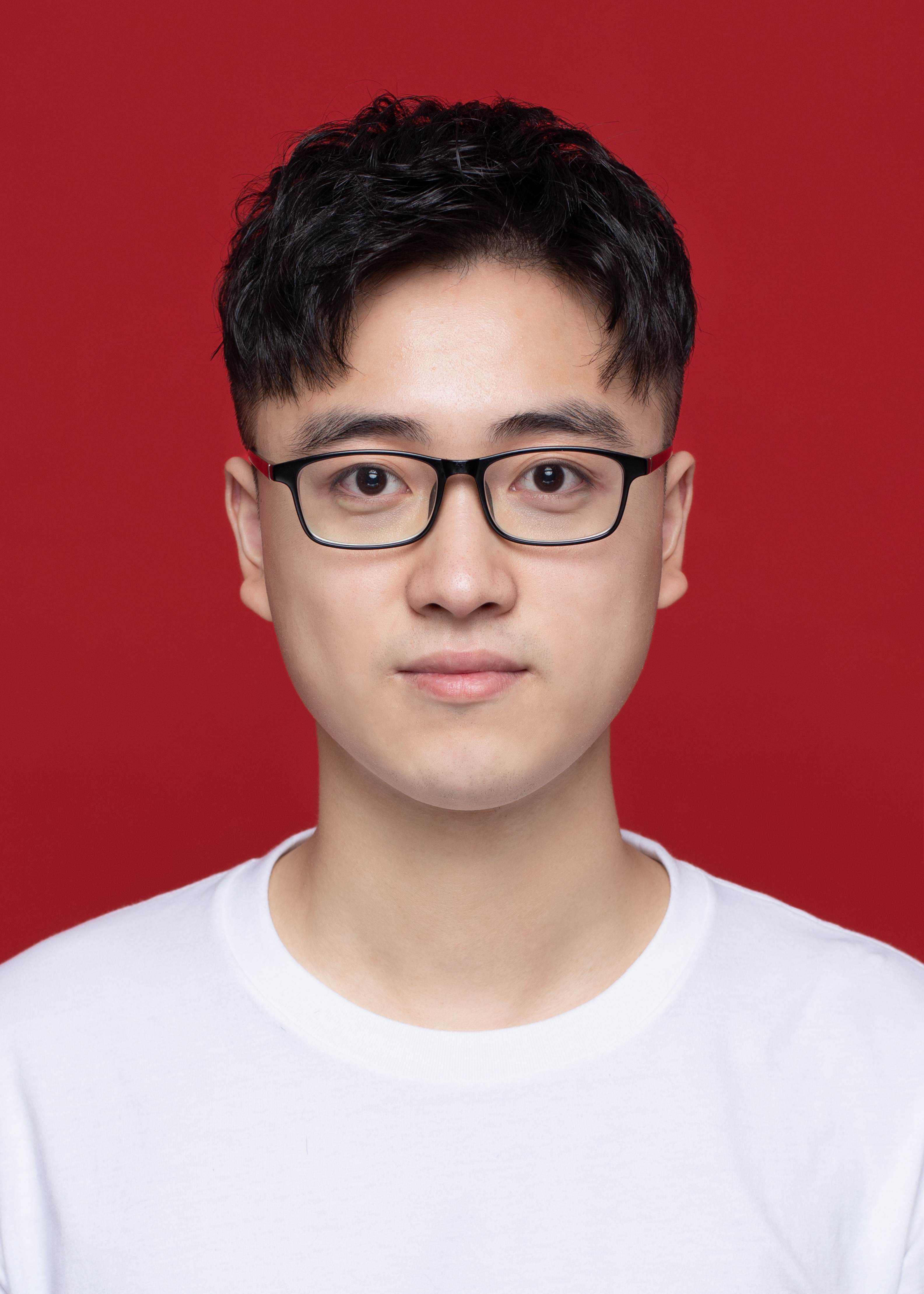}}]{Xiaochuan Dai} received his B.S. degree in 2020 from Sichuan University of Arts and Science majoring in digital media technology in China. He is currently pursuing a M.S. degree in computer technology at the College of Computer Science and Technology, Chongqing University of Posts and Telecommunications. His research interests include granular computing and data mining.
	\end{IEEEbiography}

	\begin{IEEEbiography}[{\includegraphics[width=1in,height=1.25in,clip,keepaspectratio]{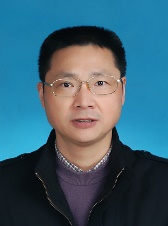}}]{Guoyin Wang} received a B.E. degree in computer software, an M.S. degree in computer software, and a Ph.D. degree in computer organization and architecture from Xi’an Jiaotong University, Xi’an, China, in 1992, 1994, and 1996, respectively. He worked at the University of North Texas, USA, and the University of Regina, Canada, as a visiting scholar during 1998–1999. Since 1996, he has been working at the Chongqing University of Posts and Telecommunications, Chongqing, China, where he is currently a professor and a Ph.D. supervisor; he is the Director of the Chongqing Key Laboratory of Computational Intelligence and the Dean of the Graduate School. His research interests include data mining, machine learning, rough sets, granular computing, cognitive computing, etc. Dr. Wang is the Steering Committee Chair of the International Rough Set Society (IRSS), a Vice-President of the Chinese Association for Artificial Intelligence (CAAI), and a council member of the China Computer Federation (CCF). He has published 300+ papers in journals and conferences, including IEEE T-PAMI, T-KDE, T-IP, T-NNLS, and T-CYB.
	\end{IEEEbiography}

	\begin{IEEEbiography}[{\includegraphics[width=1in,height=1.25in,clip,keepaspectratio]{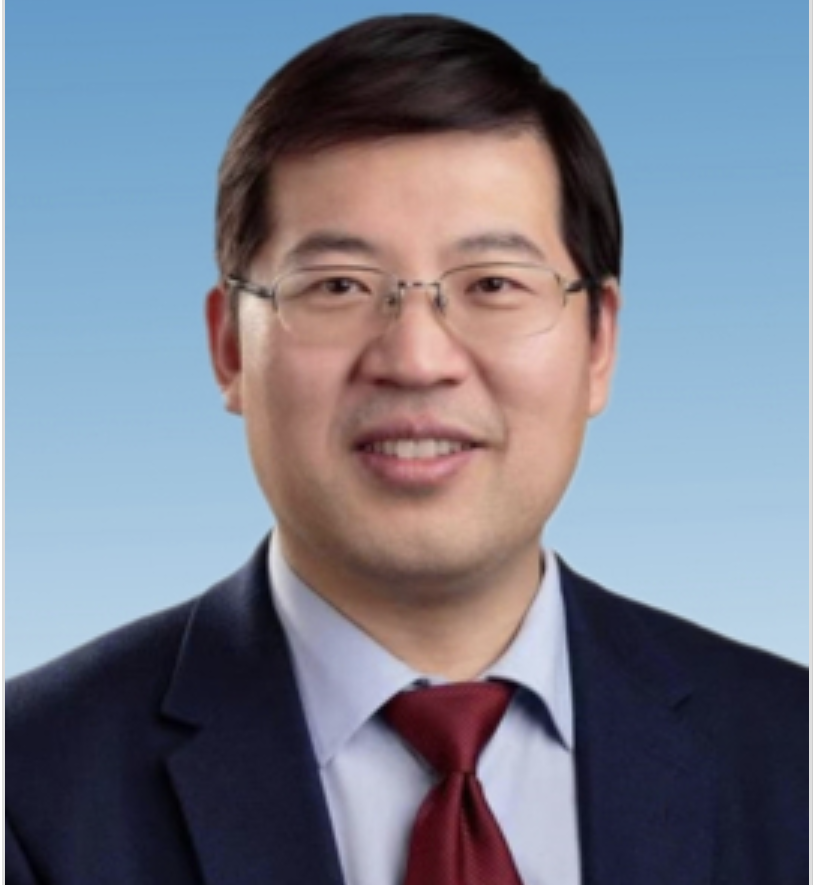}}]{Xinbo Gao} (M’02-SM’07) received BEng, MSc, and PhD degrees in signal and information processing from Xidian University, Xi’an, China, in 1994, 1997, and 1999, respectively. From 1997 to 1998, he was a research fellow with the Department of Computer Science, Shizuoka University, Shizuoka, Japan. From 2000 to 2001, he was a postdoctoral research fellow with the Department of Information Engineering, the Chinese University of Hong Kong, Hong Kong. From 2001 to 2020, he has been at the School of Electronic Engineering, Xidian University. He is currently the president of Chongqing University of Posts and Telecommunications. He has published six books and approximately 200 technical articles in prestigious journals and conferences, including IEEE T-PAMI, T-IP, T-NNLS, T-MI, NIPS, CVPR, ICCV, AAAI and IJCAI.
	\end{IEEEbiography}

	\begin{IEEEbiography}[{\includegraphics[width=1in,height=1.25in,clip,keepaspectratio]{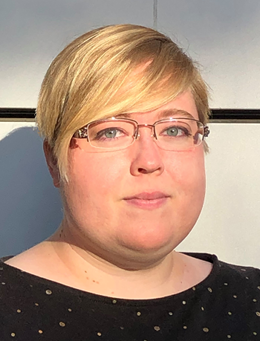}}]{Elisabeth Giem} received two bachelor's degrees, one in pure mathematics and one in music (concentration in performance) from the University of California, Riverside (UCR). She received a master's degree in computational and applied mathematics from Rice University, and a master's degree in pure mathematics from UCR. She joined Zizhong Chen's SuperLab in 2018 as a computer science Ph.D. student, and has been awarded the NICE: NRT in Integrated Computational Entomology Fellowship. Her research interests include but are not limited to high-performance computing, parallel and distributed systems, big data analytics, computational entomology, and numerical linear algebra algorithms and software.
	\end{IEEEbiography}
	
\end{document}